\documentclass[11pt,twoside]{article}

\usepackage{epsf}
\usepackage{fancyhdr}
\usepackage{graphics}
\usepackage{graphicx}
\usepackage{psfrag}
\usepackage{multirow}
\usepackage{algorithmic}
\usepackage{comment}

\usepackage[linesnumbered,ruled]{algorithm2e}
\DontPrintSemicolon

\usepackage{color}

\usepackage{amsthm}
\usepackage{amsfonts}
\usepackage{amsmath}
\usepackage{amssymb,bbm}
\usepackage{url}
\usepackage[colorlinks=True,linkcolor=magenta,citecolor=blue,urlcolor=blue,pagebackref=true,backref=true]
{hyperref}
\renewcommand*{\backref}[1]{\ifx#1\relax \else Page #1 \fi}
\renewcommand*{\backrefalt}[4]{%
    \ifcase #1 \footnotesize{(Not cited.)}%
    \or        \footnotesize{(Cited on page~#2.)}%
    \else      \footnotesize{(Cited on pages~#2.)}%
    \fi}

\usepackage{nicefrac}

\usepackage{chngpage}

 \usepackage{tabularx}%

\usepackage{enumitem}
\usepackage{booktabs}
\usepackage{pbox}

\usepackage{caption,subcaption}

\usepackage{mathtools}

\usepackage{fullpage}

\newcommand{\samples}{\ensuremath{n}}
\newcommand{\obs}{\samples}
\newcommand{\real}{\ensuremath{\mathbb{R}}}

\newcommand{\bigO}{\mathcal{O}}

\newcommand{\NORMAL}{\ensuremath{\mathcal{N}}}

\newcommand{\brackets}[1]{\left[ #1 \right]}
\newcommand{\parenth}[1]{\left( #1 \right)}

\newcommand{\abss}[1]{\left| #1 \right |}

\newcommand{\Rspace}{\ensuremath{\mathbb{R}}}

\newcommand{\mydefn}{\ensuremath{:=}}
\newcommand{\poincare}{\ensuremath{\tilde{C}}}

\newcommand{\defn}{:=}

\newcommand{\matsnorm}[2]{|\!|\!| #1 | \! | \!|_{{#2}}}
\newcommand{\vecnorm}[2]{\left\| #1\right\|_{#2}}

\newcommand{\opnorm}[1]{\ensuremath{\matsnorm{#1}{\tiny{\mbox{op}}}}}

\newcommand{\inprod}[2]{\ensuremath{\langle #1 , \, #2 \rangle}}

\newcommand{\Exs}{\ensuremath{{\mathbb{E}}}}
\newcommand{\Prob}{\ensuremath{{\mathbb{P}}}}

\newtheoremstyle{named}{}{}{\itshape}{}{\bfseries}{.}{.5em}{\thmnote{#3's }#1}
\theoremstyle{named}

\theoremstyle{plain}

\newtheorem{theorem}{Theorem}
\newtheorem{proposition}{Proposition}
\newtheorem{lemma}{Lemma}

\newlength{\widebarargwidth}
\newlength{\widebarargheight}
\newlength{\widebarargdepth}

\makeatletter
\long\def\@makecaption#1#2{
        \vskip 0.8ex
        \setbox\@tempboxa\hbox{\small {\bf #1:} #2}
        \parindent 1.5em  
        \dimen0=\hsize
        \advance\dimen0 by -3em
        \ifdim \wd\@tempboxa >\dimen0
                \hbox to \hsize{
                        \parindent 0em
                        \hfil
                        \parbox{\dimen0}{\def\baselinestretch{0.96}\small
                                {\bf #1.} #2
                                }
                        \hfil}
        \else \hbox to \hsize{\hfil \box\@tempboxa \hfil}
        \fi
        }
\makeatother

\long\def\comment#1{}
\definecolor{battleshipgrey}{rgb}{0.52, 0.52, 0.51}
\definecolor{darkgray}{rgb}{0.66, 0.66, 0.66}
\definecolor{darkgreen}{rgb}{0.0, 0.2, 0.13}
\definecolor{darkspringgreen}{rgb}{0.09, 0.45, 0.27}
\definecolor{dukeblue}{rgb}{0.0, 0.0, 0.61}
\definecolor{olivedrab7}{rgb}{0.24, 0.2, 0.12}
\definecolor{darkblue}{rgb}{0.0, 0.0, 0.55}
\definecolor{darkscarlet}{rgb}{0.34, 0.01, 0.1}
\definecolor{candyapplered}{rgb}{1.0, 0.03, 0.0}
\definecolor{ao(english)}{rgb}{0.0, 0.5, 0.0}
\definecolor{applegreen}{rgb}{0.55, 0.71, 0.0}

\newcommand{\widgraph}[2]{\includegraphics[keepaspectratio,width=#1]{#2}}

\newtheorem{assumption}{Assumption}
\newcommand{\cheeger}{\ensuremath{\zeta}}
\newcommand{\mprob}{\ensuremath{\mathbb{P}}}
\newcommand{\dtv}{\ensuremath{d_{\mbox{\tiny{TV}}}}}
\newcommand{\Event}{\mathcal{E}}

\begin{document}


\begin{center}

  {\bf{\LARGE{Sampling for Bayesian Mixture Models: MCMC with
        Polynomial-Time Mixing}}}

\vspace*{.2in}
 {\large{
 \begin{tabular}{ccc}
  Wenlong Mou$^{\diamond}$ & Nhat Ho$^{\diamond}$ & Martin
  J. Wainwright$^{\diamond, \dagger, \ddagger}$ \\
 \end{tabular}
 \begin{tabular}
 {ccc}
  Peter L. Bartlett$^{\diamond, \dagger}$ & 
  Michael I. Jordan$^{\diamond, \dagger}$
 \end{tabular}
}}

 \vspace*{.2in}

 \begin{tabular}{c}
 Department of Electrical Engineering and Computer
 Sciences$^\diamond$\\ Department of Statistics$^\dagger$ \\ UC
 Berkeley\\
 Voleon Group, Berkeley$^\ddagger$
 \end{tabular}

\vspace*{.2in}

\today

\vspace*{.2in}

\begin{abstract}
  We study the problem of sampling from the power posterior
  distribution in Bayesian Gaussian mixture models, a robust version
  of the classical posterior. This power posterior is known to be
  non-log-concave and multi-modal, which leads to exponential mixing
  times for some standard MCMC algorithms.  We introduce and study the
  Reflected Metropolis-Hastings Random Walk (RMRW) algorithm for
  sampling. For symmetric two-component Gaussian mixtures, we prove
  that its mixing time is bounded as $d^{1.5}(d + \Vert \theta_{0}
  \Vert^2)^{4.5}$ as long as the sample size $n$ is of the order $d (d
  + \Vert \theta_{0} \Vert^2)$.  Notably, this result requires no
  conditions on the separation of the two means. En route to proving
  this bound, we establish some new results of possible independent
  interest that allow for combining Poincar\'{e} inequalities for
  conditional and marginal densities.
\end{abstract}
\end{center}

\section{Introduction}

Bayesian mixture models are a popular class of models, frequently used
for the purposes of density estimation
(e.g.,~\cite{Ghosal-Ghosh-Ramamoorthi-99, Ghosal-2001, Ghosal-2007}).
Various researchers have studied posterior inference of parameters in
Bayesian mixture models~\cite{Ishwaran-2001, Nguyen-13,
  Aritra_Ho-2019}, so that the statistical behavior of such models is
relatively well-understood.  In contrast, much less is known about the
efficiency of different algorithms for sampling from the posterior
distributions that arise from Bayesian mixture models.  A standard
approach for doing so is via some form of Markov Chain Monte Carlo
(MCMC).  Many different types of MCMC algorithms have been introduced
for various types of Bayesian mixture models, including finite
Bayesian mixture models~\cite{Green-Richardson, Stephens-2000,
  Stephens-2002, Jasra-2005, Miller-2018}, Dirichlet process mixture
models~\cite{Maceachern-1998, Neal-2000, Jain-2004, Kalli-2011}, and
hierarchical and nested Dirichlet process models~\cite{Teh-et-al06,
  Rodiguez-2008}.  Despite the plethora of possible MCMC methods,
upper bounds on their mixing times are often challenging to establish.
We refer the reader to the
papers~\cite{Jones2004,Belloni2009,Woodard2013,Schreck2015,YanWaiJor16}
for non-asymptotic upper bounds on mixing times for certain types of
Bayesian models, different from those studied in this paper.

In recent years, it has been increasingly common in the Bayesian
literature to make use of a fractional likelihood---meaning an
ordinary likelihood raised to some fractional power.  Combining such a
fractional likelihood with a prior distribution in the usual way leads
to a class of posteriors known as \emph{power posterior} or
\emph{fractional posterior} distributions. The power posterior
distributions have been shown to have attractive properties in terms
of robustness to mis-specification in Bayesian mixture
models~\cite{Miller-2019}, and have been used in various applications
(e.g.,~\cite{Tavare-1997, Beaumont-2002}).  Some theoretical recent
work by Yang et al.~\cite{Yang-2019} provides contraction rates for a
range of power posteriors.

At the same time, there is now a substantial and evolving line of work
on algorithms for sampling from continuous distributions that are
based on discretizations of SDEs such as the Langevin
diffusion~\cite{RobTwe96}.  Such algorithms are now relatively
well-understood when the target distribution is log-concave, with many
provably efficient algorithms proposed (e.g., see the
papers~\cite{dwivedi2018log,cheng2017underdamped,lee2018algorithmic,mangoubi2018dimensionally,dalalyan2017theoretical}
and references therein).  The efficiency of sampling from log-concave
densities stems from their favorable isoperimetric
properties~\cite{bobkov1999isoperimetric}, which ensure that the
continuous-time Langevin diffusion mixes very fast.  By way of
contrast, sampling from multi-modal distributions is known to be hard in the worst case. There are multi-modal densities for which even the
continuous-time diffusion can take exponential time in order to escape
from a basin of attraction~\cite{bovier2004metastability}.  However,
these results are worst case in nature, and so do not preclude the
existence of efficient algorithms for particular multi-modal
densities, such as those arising from Bayesian mixture models.

Recently, a number of researchers have tackled the algorithmic
challenge of sampling from multi-modal distributions. For target
distributions that satisfy a distant dissipativity condition, it is
possible to prove global convergence, with the rate depending on
quantities such as the spectral gap, log-Sobolev constant or Stein
factor~\cite{raginsky2017non,erdogdu2018global}.  These quantities
typically lead to a mixing time that grows exponentially in the
problem parameters.  When the potential function is strongly convex
outside a ball and non-convex inside this ball, there are bounds on
the mixing time of the Langevin and Hamiltonian Monte Carlo
algorithms~\cite{cheng2018sharp,ma2018sampling,bou2018coupling}.  For
this class of algorithms, the time complexity is typically exponential
in the product $LR^2$, where $L$ is a smoothness parameter and $R$ is
the radius of that ball.

If we consider such results in the context of power posterior
distributions in Bayesian mixtures, the radius $R$ is potentially
large and dimension-dependent.  For instance, for a symmetric
two-component mixture model, with means at $\theta_0$ and $-\theta_{0}$,
the radius $R$ scales proportionally with $\|\theta_0\|_2$, a measure
of the separation between the components. Thus, a curious phenomenon
arises: although stronger separation between the components makes the
estimation problem easier, it makes the sampling problem more
difficult (at least in terms of the best known upper bounds).  Indeed,
some past work~\cite{Celeux-2000} shown that the complexity of
sampling from these posteriors can grow exponentially for certain
algorithms.

A line of recent work~\cite{lee2018beyond,ge2018simulated} has
attacked the multi-modal sampling problem by simulated tempering. In particular, these
authors studied gradient-based algorithms for sampling from mixture of
homogeneous strongly-log-concave distributions. Sampling from such
distributions can be directly solved by log-concave sampling methods;
therefore, difficulties in the simulated tempering approach mainly
arise from the restriction to gradient oracles. For power posterior sampling in mixture models, 
even if oracle access is not restricted to gradients and the structure of the landscape is known, difficulties still persist.


\subsection{Contributions}

The main contribution of this paper is to propose a particular
algorithm for sampling in a class of Bayesian mixture models, and to
prove that it has mixing time bounded by a polynomial function of
dimension and other problem parameters.  More specifically, we
consider a power posterior distribution of the form
\begin{align}
  \label{EqnMartin}
\Pi_{\obs, \beta/\obs} \Big(\theta \mid \{X_i\}_{i=1}^\obs \Big) & \;
\propto \; \; \prod \limits_{i = 1}^{\obs}
\parenth{f_{\theta}(X_{i})}^{\beta/\obs} \lambda(\theta),
\end{align}
where $\beta \in (0, n)$ is the power parameter; $\lambda$ is a prior
over $\theta$; $f_{\theta_{0}}(x) \defn \frac{1}{2} \varphi(x;
\theta_{0}, I_{d}) + \frac{1}{2}\varphi(x; - \theta_{0}, I_{d})$ is
the density of a two-component Gaussian mixture model in $\real^d$.
The class of power posterior distributions is a generalization of the
usual posterior, to which it reduces when $\beta = n$.

Of course, the posterior is a random object, since the observed data
$\{X_i\}_{i=1}^n$ have been drawn randomly from the mixture
distribution. Our main contributions are to prove certain high
probability guarantees on the behavior of an algorithm for drawing
samples from the posterior, where the probability is taken over the
randomness of the observed data.
\begin{itemize}
    \item We develop an MCMC algorithm, referred to as the
      \emph{Reflected Metropolis-Hastings Random Walk} (RMRW)
      algorithm, for drawing samples from posterior distributions of
      the form~\eqref{EqnMartin}. We prove that with high probability
      over the randomness of the samples, this algorithm has
      \emph{polynomial mixing time}---in particular, running it for
      $\mathrm{poly}(d, \Vert \theta_0 \Vert, \log
      \frac{1}{\varepsilon})$ steps yields a sample from a
      distribution that is $\varepsilon$-close to the correct
      posterior distribution in total variation (TV) distance, as long
      as the number of samples satisfies $n \gtrsim d^2 + \beta^2 d$.  The result
      does not require any separation between the components of the
      mixture distribution.  Furthermore, we demonstrate that RMRW
      algorithm also achieves polynomial mixing time under model
      mis-specification.
    \item Despite the seemingly simplicity of the symmetric Gaussian
      mixtures, the sampling problems associated with their power
      posterior are challenging.  In particular, the population and empirical log-likelihoods
      for the symmetric Gaussian mixtures have two basins of
      attraction~\cite{Siva_2017}, which are symmetric. There is a potentially high cost for
      moving across the saddle point. In order to avoid such high
      cost, we allow the RMRW algorithm to jump between components
      directly.
     \item For the population landscape, fast mixing within each
       partition requires an isoperimetric inequality.  A technical
       challenge arises in establishing this inequality: despite the
       unimodality of each basin of attraction, they are only
       \emph{quasi-concave}, and such distributions are known to have
       poor isoperimetry in the worst
       case~\cite{chandrasekaran2009sampling}. Moreover, there can be
       a non-trivial proportion of mass assigned to the neighborhood
       of the saddle point.  For this reason, the Markov chain has to
       explore the part with negative curvature in a careful way,
       instead of entirely avoiding it by a suitably large
       initialization~\cite{Hsu-nips2016}.  Addressing these
       properties requires a careful analysis of the geometry of
       symmetric Gaussian mixtures.  In particular, the population
       log-likelihood of these models is quasi-convex in the direction
       of $\theta_{0}$ restricted to a partition, and convex in all
       other directions. In order to show isoperimetry for this
       function and guarantee good isoperimetry within each partition,
       we establish general structural results that combine the
       Poincar\'{e} inequalities for conditional and marginal
       densities. Combining these results leads to the polynomial-time
       mixing rate guarantee for the RMRW algorithm.
\end{itemize}

We note in passing that our novel Poincar\'{e} inequality, while
applied here only to Gaussian mixtures, is of potential use for a much
broader class of non-log-concave densities. Our isoperimetric
inequalities are based only on geometric properties, so can hold for
distributions that need not be log-concave.  The bounds from past work
on sampling from mixture
distribution~\cite{lee2018beyond,ge2018simulated} are based directly
on the isoperimetric constants for each mixture component, and so
require each mixture component to be well-behaved.

By contrast, the isoperimetric inequalities proved in this paper are
linked directly to the geometric structure of the target density.  For
this reason, we suspect that our results may be useful for analyzing
sampling algorithms for more general classes of distributions.
Finally, our proof follows the avenue of relating the sample
log-likelihood to the population log-likelihood via empirical process
theory, and then exploiting the structure of the population problem.
This proof technique is also applicable to sampling problems from
other types of Bayesian posterior distributions.

In the existing literature, many results have been
established to derive log-Sobolev inequalities by combining those for
marginals and
conditionals~\cite{otto2007new,lelievre2009general,grunewald2009two}.
Such results play a key role in understanding the dynamics of certain
statistical physics systems. Our technical lemma for combining the
Poincar\'{e} inequalities can be seen as parallel to this existing
literature. In our setup, it is also important that only the Poincar\'{e}
inequalities are assumed for the marginal and conditional
distributions; indeed, the joint distribution is not
strongly-log-concave and may not satisfy log-Sobolev inequality with a good constant.


\subsection{Organization and notation}

The remainder of the paper is organized as follows.  In
Section~\ref{sec:setup}, we provide the setup for symmetric Gaussian
mixtures and several useful definitions for establishing mixing time
of MCMC algorithms.  In Section~\ref{sec:population}, we describe the
RMRW algorithm including some of the underlying geometry, and sketch
out our analysis of its mixing time.  Several key results with the
isoperimetric inequalities and conductance of Markov chain defined by
RMRW algorithm are presented in Section~\ref{sec:geometry} and
Section~\ref{sec:conductance}.  The proofs for key results in the
paper are in Section~\ref{sec:proofs} whereas the proofs for remaining
results are deferred to the appendices.  We conclude with a discussion
in Section~\ref{sec:discussion}.

\paragraph{Notation and basic definitions.} For each positive integer $n$,
we use $[n]$ to denote the set $\{1, 2, \ldots, n\}$.  For any set
$A$, we denote $A^{c}$ as its complement.  For any vector $x
\in \Rspace^{d}$, we denote $x_{i}$ as its $i$-th component while
$x_{-i}$ stands for all the components except $i$-th component for $i
= 1, \ldots, d$. The expression $a_{n} \succsim b_{n}$ will be used to
denote $a_{n} \geq c b_{n}$ for some positive universal constant
$c$. Given two density functions $p$ and $q$ with respect to Lebesgue
measure $\mu$, the squared Hellinger distance between $p$ and $q$ is
given by $h^{2}(p, q) = \frac{1}{2} \int (\sqrt{p(x)} - \sqrt{q(x)})^2
d\mu(x)$.  The total variation (TV) distance and Kullback-Leibler (KL)
divergence are given by
\begin{align*}
\dtv(p, q) = \frac{1}{2} \int \abss{p(x) - q(x)} d\mu(x), \quad
\mbox{and} \quad \text{KL}(p, q) = \int \log(p(x)/ q(x)) p(x) d\mu(x),
\end{align*}
respectively. A function $f: \Rspace^d \to \Rspace$ is \emph{quasi-convex} if for
any $\alpha \in \Rspace$, the level set $\{ x \in \real^d \, \mid \,
f(x) \leq \alpha \}$ is a convex set.


\section{Problem set-up}
\label{sec:setup}

In this paper, we study mixing time for the Bayesian posterior
distributions that are induced by a symmetric two-component location
Gaussian mixture model, or in short, a \emph{symmetric Gaussian
  mixture}. An instance of such a mixture model is characterized by a
density of the form
\begin{align}
\label{eq:symmetric_Gaussian}
f_{\theta_{0}}(x) & \mydefn \frac{1}{2} \varphi(x; \theta_{0}, I_{d})
+ \frac{1}{2}\varphi(x; - \theta_{0}, I_{d}).
\end{align}
Here $\varphi( \: \cdot \: ; \theta, I_{d})$ denotes the multivariate
Gaussian distribution with location parameter $\theta \in \Rspace^{d}$
and covariance matrix $I_{d}$. We are interested in the problem of
sampling from the \emph{power posterior distribution} induced by a
symmetric Gaussian mixture and a prior $\lambda$.  It takes the form
\begin{align}
\label{eq:power_posterior}
\Pi_{\obs, \beta/\obs} \Big(\theta \mid \{X_i\}_{i=1}^\obs \Big)
\mydefn \dfrac{\prod \limits_{i = 1}^\obs
  \parenth{f_{\theta}(X_{i})}^{\beta/ \obs}
  \lambda(\theta)}{{\displaystyle \int \prod \limits_{i = 1}^{n}
    \parenth{f_{u}(X_{i})}^{\beta/\obs} \lambda(u) d u}},
\end{align} 
where $\beta \in (0, \obs)$ is a parameter.  While the setting of
$\beta$ typically varies as a function of the sample size $\obs$, here
we omit this dependence so as to simplify notation. The exponentiated
likelihood in the numerator (disregarding the prior term) is often
referred to as the \emph{power likelihood}.

Note that sampling from the distribution~\eqref{eq:power_posterior} is
equivalent to sampling from the distribution with density $\pi$
proportional to $e^{-U( \theta)}$, where
\begin{align}
\label{eq-posterior}
U( \theta) \mydefn \frac{ \beta}{n} \sum_{i = 1}^n \log \left(
\frac{1}{2} \varphi( \theta - X_i) + \frac{1}{2} \varphi(\theta + X_i)
\right) + \log \lambda( \theta).
\end{align}
For the symmetric Gaussian mixtures, the power posterior distribution
is non-log-concave and multi-modal.  In such context, there are few
polynomial-time guarantees on sampling algorithms.  Most existing
works on sampling from multi-modal distributions require the function
$-\log \pi$ to be convex outside a ball, and have exponential
dependence on the radius of this
ball~\cite{cheng2018sharp,ma2018sampling,bou2018coupling}.  When
applying these results to power posterior sampling in symmetric
Gaussian mixtures, the radius of this ball scales with separation
between mixture components, i.e., $\Vert \theta_{0} \Vert$.  From this
fact arises a curious phenomenon: while larger separations between
mixture components makes estimation easier, it appears to make
sampling harder, at least in terms of known upper bounds.

Note that the power posterior~\eqref{eq:power_posterior} is a random
measure, since it depends on the observations
$\{X_i\}_{i=1}^\obs$. For the purposes of analysis, it turns out to be
convenient to introduce a deterministic measure that we refer to as
the \emph{population power posterior}.  In particular, given the
symmetric Gaussian mixture~\eqref{eq:symmetric_Gaussian}, the
population power posterior is defined by a density of the form
$\pi_0(\theta)\propto e^{-U_0(\theta)}$, where
\begin{align*}
  U_0( \theta) \mydefn \beta \cdot \Exs \log \left( \frac{1}{2}
  \varphi( \theta - X) + \frac{1}{2} \varphi(\theta + X) \right) +
  \log \lambda( \theta).
\end{align*}
Here the expectation is taken over a random observation $X \sim
\frac{1}{2} \mathcal{N}( \theta_0, I_{d}) + \frac{1}{2} \mathcal{N}(-
\theta_0, I_{d})$ taken from the underlying two-component Gaussian mixture.


\section{Polynomial time MCMC algorithm}
\label{sec:population}

With these definitions in hand, we are now ready to describe our MCMC
algorithm for sampling from the power posterior~\eqref{eq:power_posterior}.  For reasons to
become clear, we refer to it as the \emph{Reflected
  Metropolis-Hastings Random Walk} (RMRW) algorithm.  We first
describe that algorithm based on population power posterior sampling
and state a formal result with its polynomial mixing time in
Section~\ref{sec:MCMC_algorithm}.  We then provide geometric intuition
behind the polynomial mixing time of RMRW algorithm, along with a
sketch that outlines some key ideas in the proof.


\subsection{Reflected Metropolis-Hastings Random Walk algorithm}
\label{sec:MCMC_algorithm}

The RMRW algorithm we develop in this section relies on the special
structure of symmetric Gaussian mixtures.  In pseudocode, it takes the
following form:
\begin{algorithm}[htb]
  \caption{Reflected Metropolis-Hastings Random Walk Algorithm}
  \label{alg-MCMC-sym}
\begin{algorithmic}
\REQUIRE Oracle access to potential function $F$, step size $\eta>0$.
\ENSURE Approximate sample $\theta\sim \pi\propto e^{-F}$.  \STATE
Sample $\theta^{(0)}\sim \mathcal{N}(0,I_{d})$.  \FOR{$t=1,2,\cdots$}
\STATE Let $Y\sim \theta^{(t-1)}+\sqrt{\eta}\mathcal{N}(0,I_{d})$.  \STATE
Let $Z=Y$ with probability $\frac{1}{2}$ and $Z=-Y$ with probability
$\frac{1}{2}$.  
\STATE Let $\theta^{(t)} = \begin{cases} Z & \text{with probability }
\min\left(1,\exp(F(Z)-F(\theta^{(t-1)}))\right)\\
\theta^{(t-1)}&\text{otherwise}
\end{cases}.$
\ENDFOR
\end{algorithmic}
\end{algorithm}

As seen from this pseudocode, the RMRW algorithm is a random walk, based
on an isotropic Gaussian proposal distribution, combined with
a reflection through origin, along with a final adjustment using a
Metropolis-Hastings correction.  Note that the algorithm does not
exploit any information about the gradient of the potential function,
and so can be seen as a zero$^{th}$-order procedure (meaning that it
only uses the value of the potential function as opposed to its gradient). 

\begin{figure}[!htb]
  \begin{center}
    \begin{tabular}{cc}
      \widgraph{0.48\textwidth}{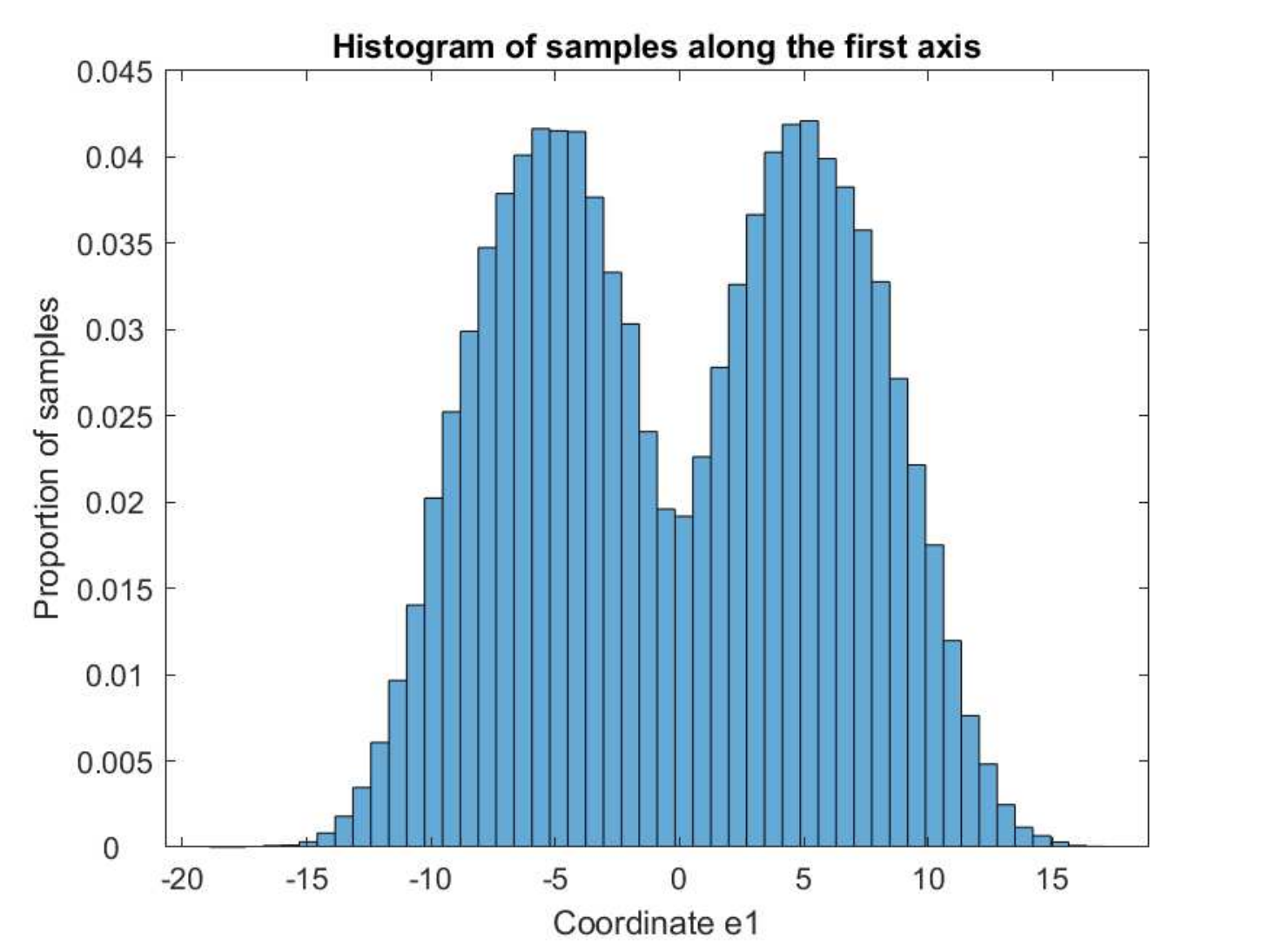} &
          \widgraph{0.48\textwidth}{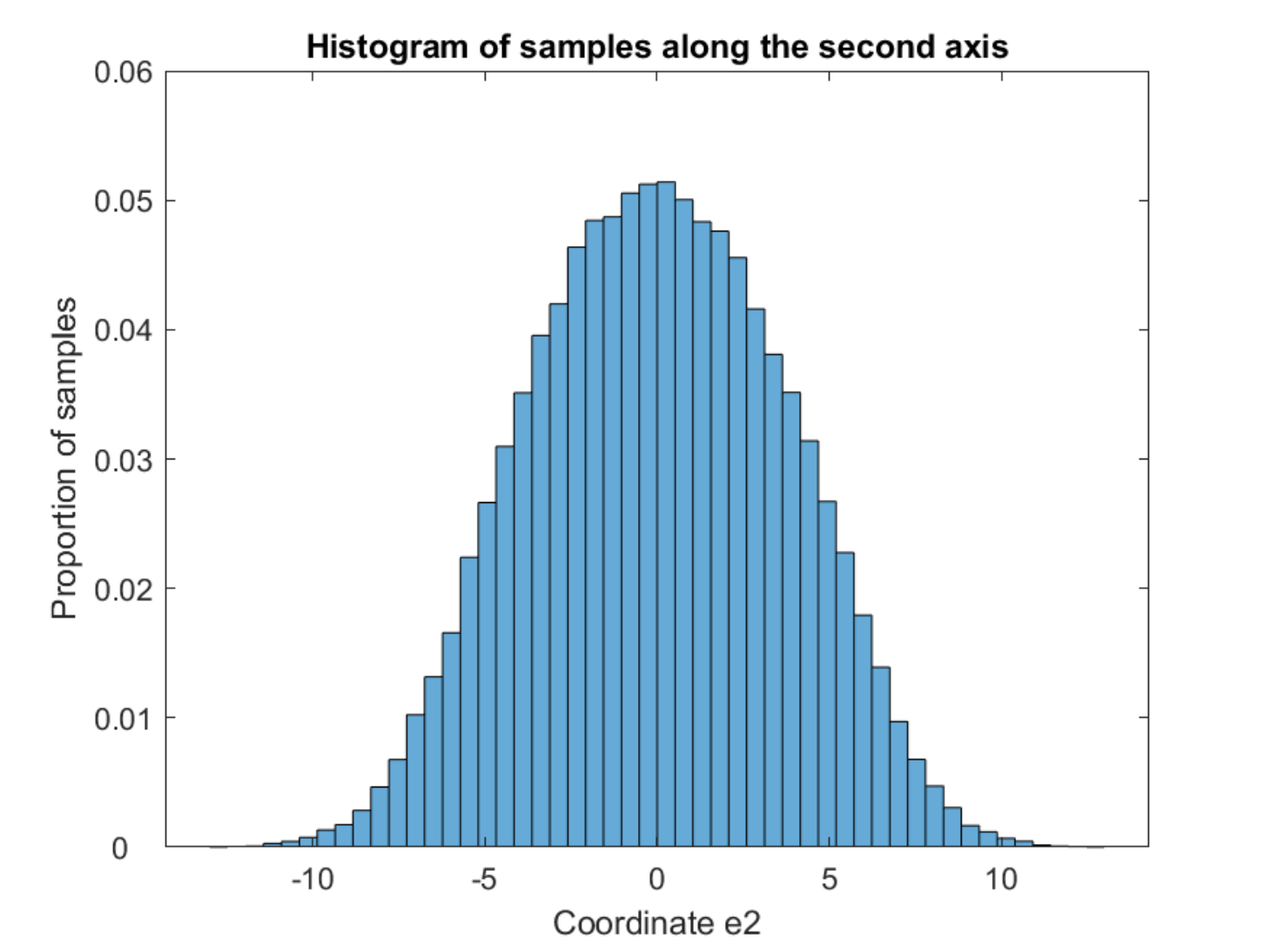}\\
          (a) & (b)
    \end{tabular}
    \begin{tabular}{cc}
      \widgraph{0.48\textwidth}{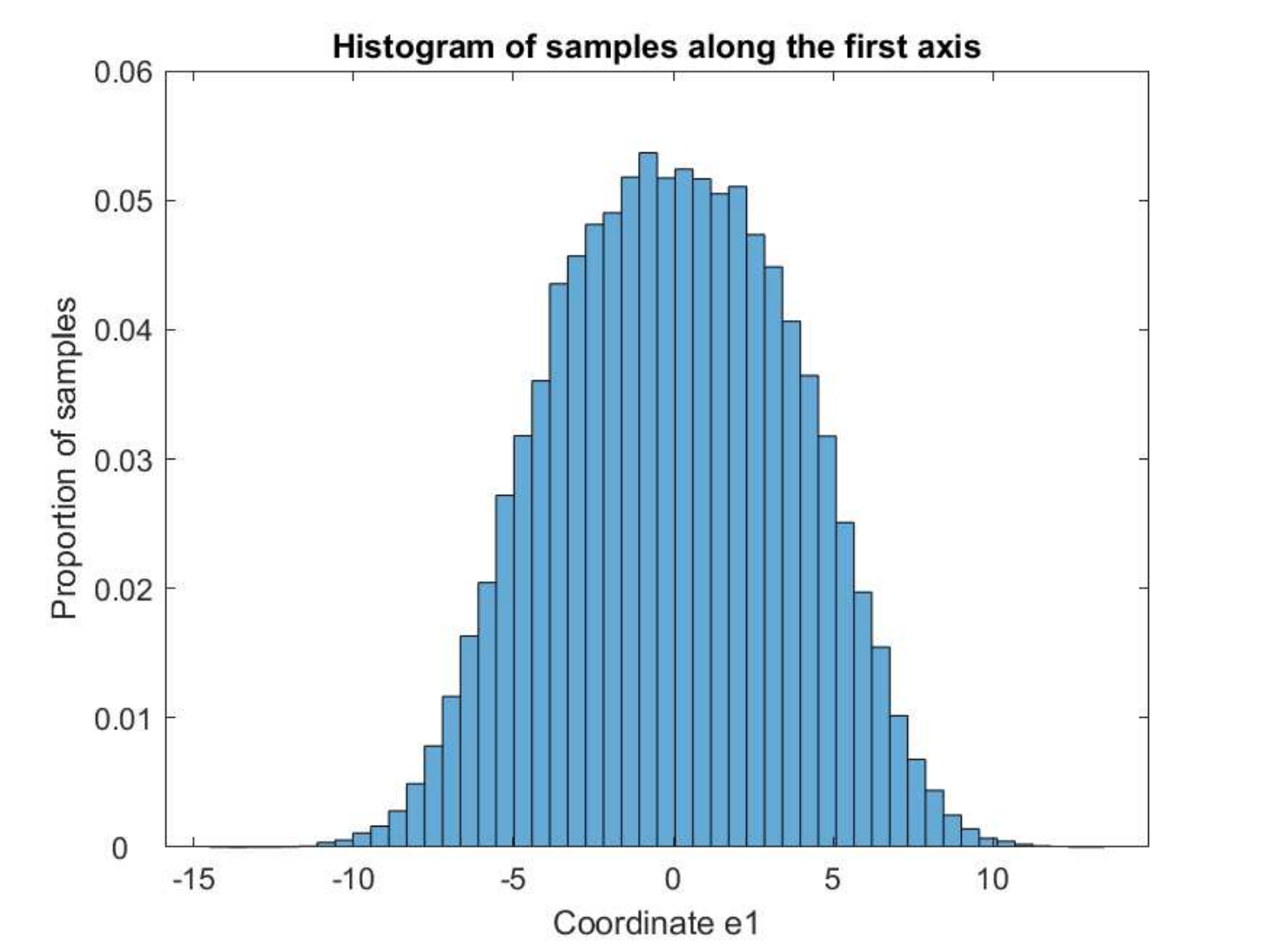} &
          \widgraph{0.48\textwidth}{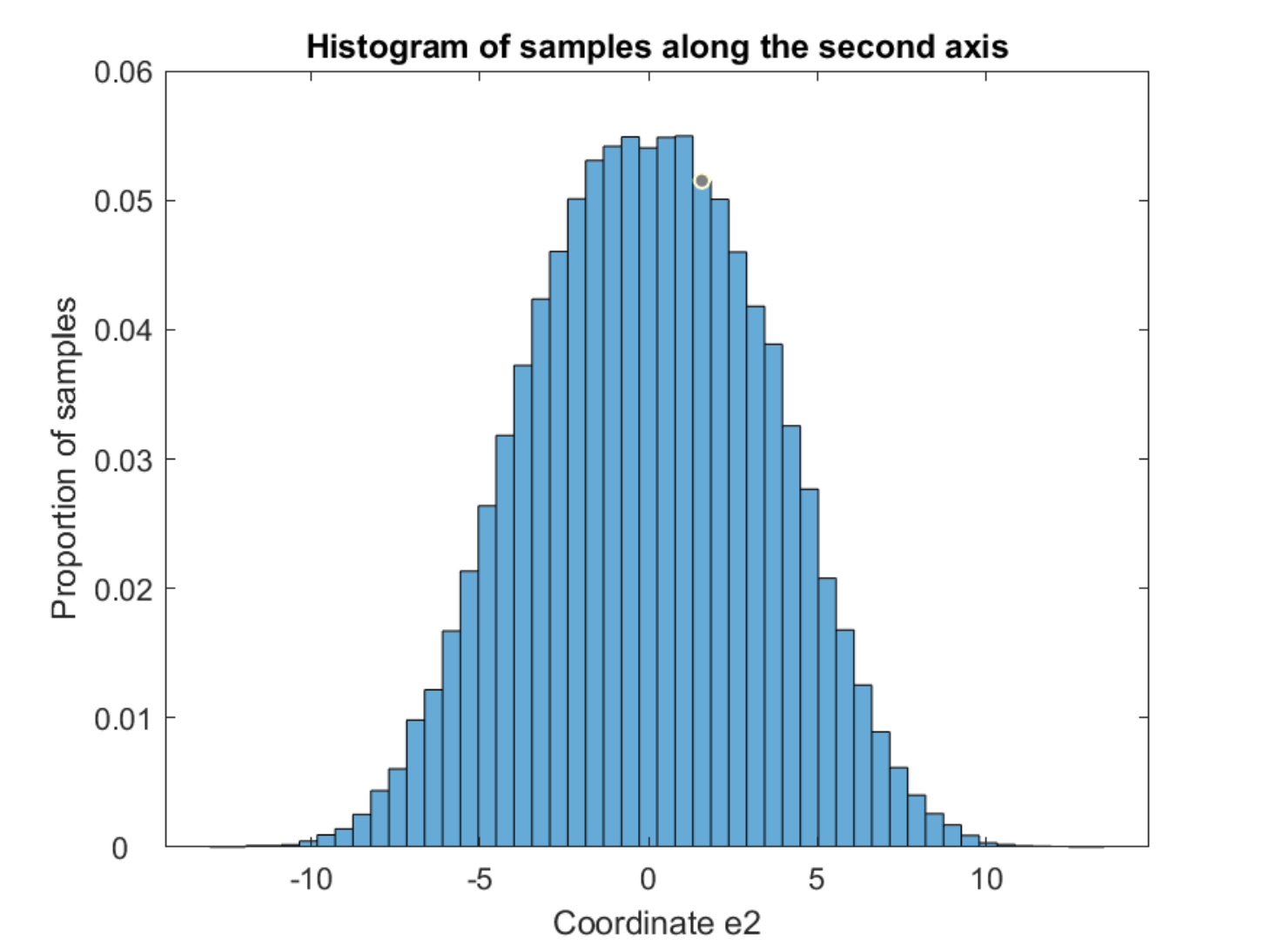}\\
          (c)  & (d) 
    \end{tabular}
    \caption{Illustration of samples from RMRW algorithm for
      approximating power posterior of a $10$-dimensional symmetric
      two-component Gaussian mixture model based on $100$ data points,
      with $\beta = 8$.  (a), (b): Case $a = 5$ for which the
      components are well-separated.  Panel (a) shows projection of
      samples onto $e_1$, the direction of the true parameter
      $\theta_{0}$, whereas plot (b) shows the projection of the
      samples onto $e_2$, which is orthogonal to the true parameter.
      Due to the separation, the projection in the direction $e_1$
      exhibits bimodality. (c), (d).  Same plots with $a = 0$, for
      which there is no separation between components.  Notice the
      biomodality has now disappeared.}
    \label{figure-RMRW_samples}
  \end{center}
\end{figure}

We also demonstrate the shape of the power posterior distribution by simulation results based on our RMRW algorithm. We consider $n = 100$ data points generated 
from a $10$-dimensional symmetric two-component Gaussian mixture model~\eqref{eq:symmetric_Gaussian} 
with $\theta_{0} = a e_1$ for $a \in \{ 0, 5 \}$ 
and $e_{1} = (1, 0, \ldots, 0)$. The fraction parameter $\beta$ is chosen as $\beta = 8$. 
We generate the samples by taking $100000$ consecutive iterates of the RMRW algorithm. 
In Figure~\ref{figure-RMRW_samples}, we show the histograms 
for the projection of samples drawn from the RMRW algorithm onto the first and second dimension.

\subsection{Bound on mixing time and geometric intuition}

It
is straightforward to show that the target distribution is a
stationary distribution of Markov chain defined by
Algorithm~\ref{alg-MCMC-sym}.  Of interest to us is a bound on the
mixing time of the algorithm, defined as
\begin{align}
\label{EqnMixing}
T(\epsilon) & \mydefn \inf_{t = 1, 2, \ldots} \left \{ d_{TV}(\pi^t,
\pi) \leq \epsilon \right \},
\end{align}
where $\pi^t$ denotes the distribution of the algorithm's iterates at
time $t$. In this paper, we study the case of a uniform and
hence improper prior---namely, the choice $\lambda(\theta) = 1$ for
all $\theta \in \real^d$.  This improper prior can be viewed as the
limiting case of the normal prior $\NORMAL(0, \sigma^2 I_{d})$ as
$\sigma \to \infty$. We note also that the techniques in the paper can
be extended to the case when $\lambda(\theta)$ is a Gaussian density
function of the form $\NORMAL(0, \sigma^2 I_{d})$ for $\sigma > 0$,
but we refrain from doing so in order to simplify the statement.  In
the following statement, we use $C_1$ and $C_2$ to denote universal
positive constants.

\begin{theorem}[Mixing time of RMRW algorithm]
  \label{theorem:mixing_time}
  Let $\varepsilon \in (0,1)$ be a target TV distance error, and let
  $\delta \in (0,1)$ be a pre-specified failure probability.  Suppose
  that the sample size $\obs$ is lower bounded as $\obs \geq C_1
  d\left(\beta^2(1+\Vert\theta_0\Vert^2)+d\right) \log(d (\Vert
  \theta_0 \Vert + 1)/ \delta)$, and the RMRW algorithm is run with
  the (random) potential function $U$ in
  equation~\eqref{eq-posterior}.  Then with probability at least
  $1-\delta$ over this randomness, the mixing time is bounded as
\begin{align*}
  T(\epsilon) & \leq C_2 \; d^{3/2} \; \left(d + \Vert \theta_0
  \Vert^2 \right)^{9/2} \log^{11/2} \left(\frac{1}{\varepsilon}
  \right).
\end{align*}
\end{theorem}
\noindent In order to provide intuition for the result in
Theorem~\ref{theorem:mixing_time}, we sketch out the main steps of the
proof in Section~\ref{sec:proof-sketch}; see
Section~\ref{sec:proof-final} for the full proof with all technical
details.


\subsection{Geometric intuition}
\label{sec:geometric-intuition}

\begin{figure}[t]
  \begin{center}
    \begin{tabular}{cc}
      \widgraph{0.45\textwidth}{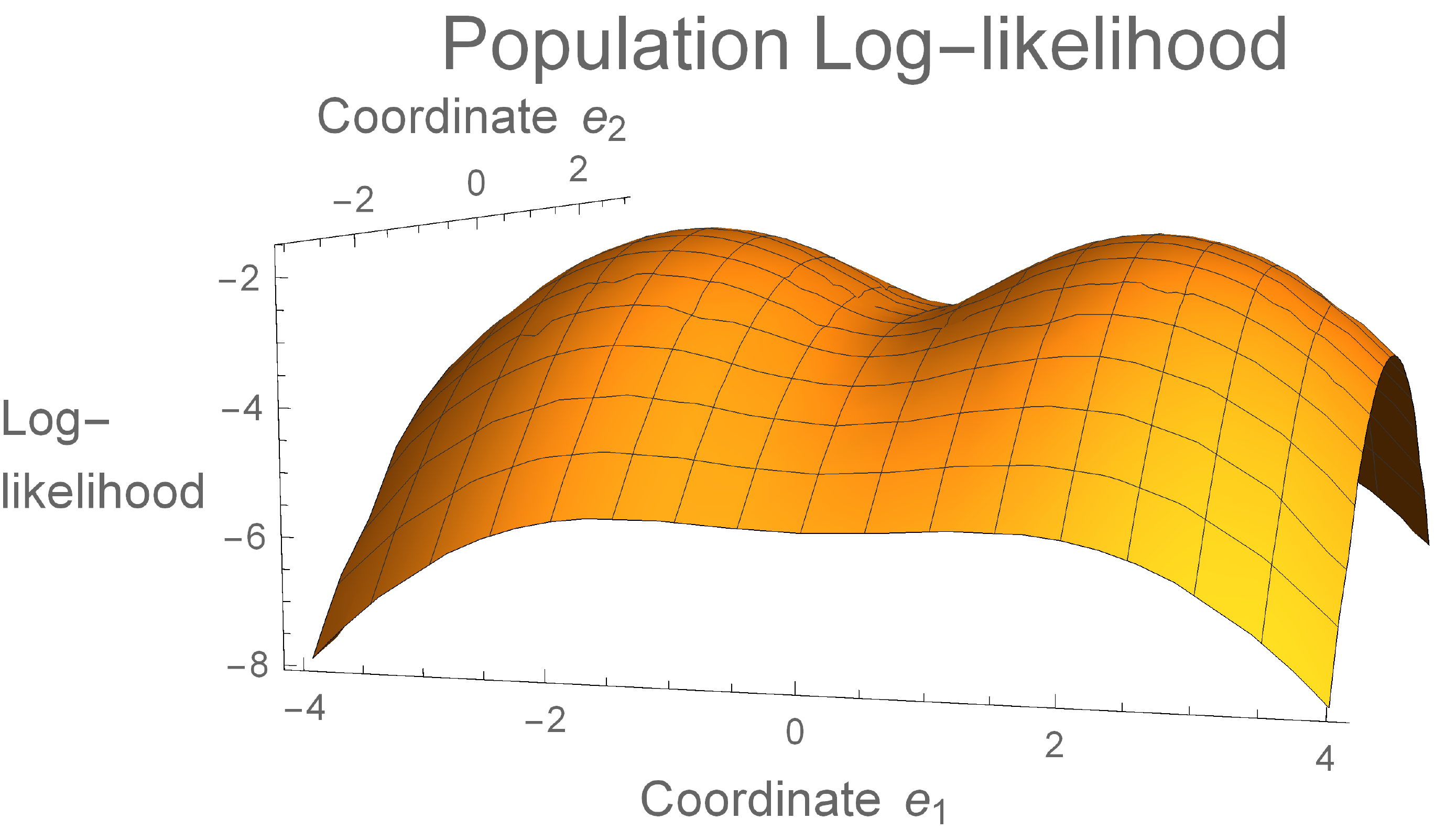} &
      \widgraph{0.45 \textwidth}{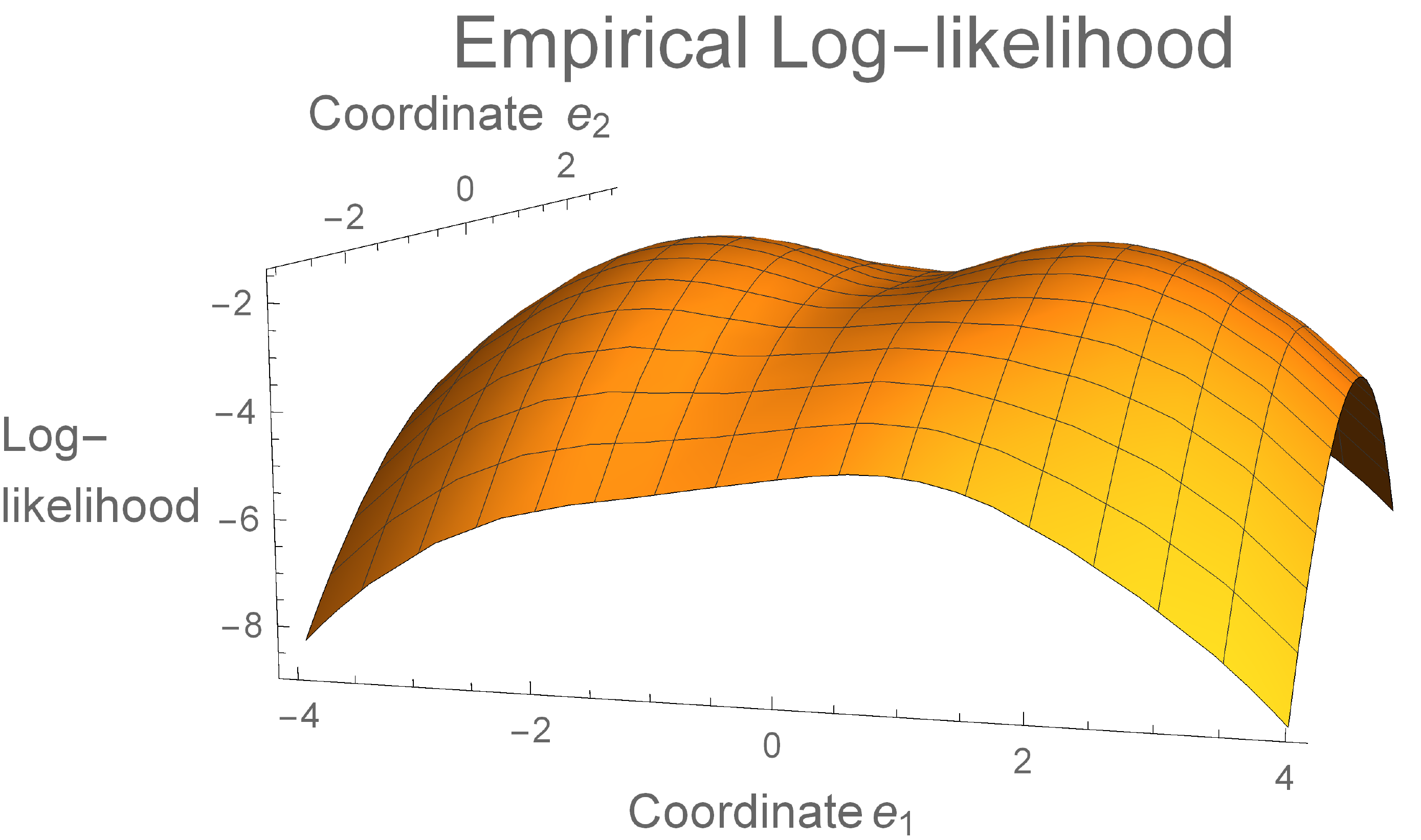}\\
      (a) & (b)
    \end{tabular}    
    \caption{(a) Log-likelihood at the population level for a Gaussian
      mixture model in dimension $d = 2$, mean $\theta_{0} = (2 , 0
      )$, and power parameter $\beta = 1$. It is symmetric around the
      origin and has two basins of attraction $\theta_0$ and $-
      \theta_0$.  (b) Log-likelihood at the empirical level, based on
      $\obs = 10$ samples. It is symmetric and has two modes with high
      probability. However, the partition between the two modes is not
      aligned with the population counterpart.}
    \label{figure-population-empirical}
  \end{center}
\end{figure}

Figure~\ref{figure-population-empirical} provides some geometric
intuition for the algorithm, and its relatively fast mixing.  Panel
(a) provides a surface plot of the negative log-likelihood at
population level for a problem in $d = 2$ dimensions.  Note that the
negative log-likelihood is symmetric around the origin and has global
minima at $\theta_0$ and $-\theta_0$.  In any direction orthogonal to
$\theta_0$, the negative log-likelihood is a convex function. The
$\theta_0$ direction is the interesting one: in this direction, the
landscape can be partitioned into two symmetric components; within
each of them, the negative log-likelihood is quasi-convex.  We exploit
these properties in order to bound the conductance of Markov chain in
RMRW algorithm.

An additional challenge is the existence of a saddle point at $\theta
= 0$.  In the context of computing point estimates for mixture models,
this difficulty can be side-stepped by initializing an iterative
algorithm with any vector of sufficiently large
norm~\cite{Hsu-nips2016}.  The sampling problem---in contrast to the
problem of point estimation---requires a complete exploration of the
probability surface, and the mass with negative curvature cannot be
ignored. Accordingly, the following sections are devoted to the
development of some new tools for establishing Poincar\'{e} and
isoperimetric inequalities over a partition. We would like to note
that though several geometric conditions can be verified analytically
for the population power posterior, the calculations are delicate and
do not generalize to the empirical counterpart. Fortunately, by the
Holley-Stroock principle, conductance results are robust with
multiplicative perturbation of the density.  As a consequence, we can
use empirical process theory to guarantee global uniform convergence
of the potential function.  This control allows us to transfer
conductance bounds from the population to the empirical power
posterior.

\subsection{Proof sketch}
\label{sec:proof-sketch}

In this section, we provide a sketch of the proof of
Theorem~\ref{theorem:mixing_time}.  The argument makes use of the
standard notions of a Poincar\'{e} inequality, and the
$s$-conductance~\cite{Lovasz-1993}, which we now define.

\paragraph{Poincar\'{e} inequality:} Given a set $\Omega \subseteq \Rspace^d$
with smooth boundary, consider a probability distribution $\pi$
supported on $\Omega$.  It is said to satisfy the \emph{Poincar\'{e}
  inequality} with constant $C > 0$ if the inequality
\begin{align*}
  \int_\Omega f^2(x) d\pi(x) \leq C \int_\Omega \Vert 
  \nabla f(x) \Vert^2 d\pi(x)
\end{align*}
holds for all functions $f$ in the Sobolev space $H^1(\pi, \Omega)$ such that
$\Exs_{\pi} f(X) = 0$.


\paragraph{$s$-conductance:}
For a space $\Omega$ accompanied with a $\sigma$-algebra $\mathcal{F}$,
we consider a discrete-time Markov chain with transition kernel
$\mathcal{T}_x(S):\Omega\times \mathcal{F}\rightarrow \Rspace$ whose
stationary distribution is $\pi$. For any $s > 0$, the $s$-conductance
of this Markov chain is
\begin{align}
\label{EqnDefnSConductance}  
  \Phi_s( \pi) \mydefn \inf_{S \in \mathcal{F}: s < \pi(S) \leq
    \frac{1}{2}} \frac{\int T_x(S^{c}) d\pi(x)}{\pi(S) - s}.
\end{align}

\noindent With these definitions in hand, we now sketch out the proof,
which consists of three main steps:

\paragraph{Step 1:} First, since the two basins of attraction in
the population log-likelihood are symmetric, we can partition the
space into two symmetric components along direction $\theta_0$, and
analyze the isoperimetry within each of them.  Even within a
partition, the density is neither log-concave nor $s$-concave, making
existing isoperimetric bounds inapplicable. However, a key geometric
observation is that the marginal density at $\theta_0$ direction is
quasi-concave, and the conditional density for other directions are
log-concave.  We derive a novel result for combining Poincar\'{e}
inequalities of these densities
(Lemma~\ref{lemma-poincare-decomposition} in
Section~\ref{sec:isoperimtry-structural}), and use it to obtain the
following isoperimetric result:
\begin{theorem}
\label{theorem-isoperimetry-partition}
  For any positive constants $A > \max(\Vert \theta_0 \Vert, 1)$ and
  $M > 1$, the Cheeger constant $\cheeger$ for density $\pi_0$
  restricted to $[0, A] \times \mathcal{B}(0, M)$ is lower bounded as
  \begin{align}
  \label{EqnIsoPartition}  
    \cheeger \geq \frac{c}{ A^5 M^2} \; \frac{1}{\sqrt{d}}.
  \end{align}
\end{theorem}
\noindent In the lower bound~\eqref{EqnIsoPartition}, the constant $c
> 0$ is independent of $A$ and $M$.  The key results that underlie
this theorem are given in Sections~\ref{sec:isoperimtry-structural}
and~\ref{subsec:iso_pop}, whereas the technical details of the proof
are given in
Section~\ref{subsec:proof:theorem-isoperimetry-partition}.

\paragraph{Step 2:} The second step is to establish conductance results for the Markov
chain defined by RMRW algorithm.
Theorem~\ref{theorem-isoperimetry-partition} provides a lower bound on
the Cheeger constants inside each partition, which implies the fast
mixing of the Metropolized random walk within each component.  In
order to make the Markov chain mix fast in the whole space without
waiting for the exponential exit time of a basin-of-attraction, we
allow it to jump between two components.  Due to the symmetry of
population log-likelihood, this leads to $s$-conductance lower bound
on the Markov chain defined by RMRW algorithm with density $\pi_0
\propto e^{- U_0}$ as follows:
\begin{align}
\label{eq-conductance-population}  
  \Phi_s( \pi_0) & \gtrsim \; d^{- 1/4} \Big \{ \sqrt{d} + \Vert
  \theta_0 \Vert + \sqrt{\log(1/s)} \Big \}^{- 9/ 2}.
\end{align}
Section~\ref{subsection:population-conductance} provides various
results that lead to this conductance bound.


\paragraph{Step 3:}  Our final step is to translate results for population
power posterior to its empirical counterpart, in particular using
techniques from empirical process theory.
Lemma~\ref{lemma-empirical-process} provides control on the deviations
between the empirical and population posteriors, up to a
multiplicative factor.  Since the $s$-conductance is robust with
respect to such multiplicative perturbations, this control allows us
to translate bounds on the mixing rate of a chain on the population
posterior to the actual Markov chain that evolves according to the
empirical power posterior.


\section{Geometric results and isoperimetry}
\label{sec:geometry}

This section is dedicated to the proof of
Theorem~\ref{theorem-isoperimetry-partition}.  We first provide
general structural results about Poincar\'{e} inequalities for certain
classes of non-log-concave densities.  We then validate the
assumptions in the case of population power posterior for the
symmetric two-component Gaussian mixture models~\eqref{eq:symmetric_Gaussian}.
Lemma~\ref{lemma-poincare-decomposition} for combining Poincar\'{e}
inequalities for conditional and marginal densities is of possible independent interest.


\subsection{General structural results}
\label{sec:isoperimtry-structural}

We establish structural lemmas that are needed to prove the
isoperimetric inequalities for the population power posterior.  The
potential function $U_0$ is convex in all directions that are
orthogonal to $\theta_0$.  In contrast, it is non-convex in the
direction $\theta_{0}$, and so a special treatment is required.  In
particular, we make use of the following lemma, which allows us to
combine Poincar\'{e} inequalities for conditional and marginal
densities together so as to obtain the Poincar\'{e} inequality for the
whole density.
\begin{lemma}
  \label{lemma-poincare-decomposition}
  Consider a probability measure $\pi$ that is continuously
  differentiable over its support $\Omega_1 \times \Omega_2$, where
  $\Omega_1 \subseteq \Rspace^{d_1}$ and $\Omega_2
  \subseteq \Rspace^{d_2}$ are closed sets. For $X = (X_1,X_2) \sim
  \pi$, suppose that
  \begin{itemize}
      \item The marginal distribution of $X_1$ satisfies a
        Poincar\'{e} inequality with constant $C_1$.
    \item For any $x_1 \in \Omega_1$, the conditional distribution
      $X_2| X_1 = x_1$ satisfies a Poincar\'{e} inequality with
      constant $C_2$ uniform in $x_1$.
  \item The function $x_1 \mapsto \log \pi (x_2 | x_1)$ is uniformly
    differentiable, and moreover, there exists a constant $L > 0$ such
    that $\sup \limits_{x_1 \in \Omega_1,x_2 \in \Omega_2}
    \vecnorm{\nabla_{x_1} \log \pi (x_2 | x_1)}{2} \leq L$.
  \end{itemize}
Under these conditions, the density $\pi$ satisfies a Poincar\'{e}
inequality with the constant
  \begin{align*}
    \poincare = 2 \left(C_1 + C_2 + C_1 C_2 L^2 \right).
  \end{align*}
\end{lemma}
\noindent
See Section~\ref{subsec:proof:lemma-poincare-decomposition} for the
proof of this claim.

Let us sketch the main ideas underlying this lemma. For any zero-mean
function $h$ on $\Omega_1 \times \Omega_2$, we need to control the
variance of $h(X_1, X_2)$ using the second moment of its gradient.
For any $x_1 \in \Omega_1$ fixed, we perform bias-variance
decomposition on $h^2(x_1, X_2)$ with $X_2 \sim \pi(x_2| x_1)$.  The
variance term can be easily dealt with by the Poincar\'{e} inequality
for the conditional density.  For the bias term $\Exs_{\pi}\left(
h(X_1, X_2)| X_1 = x_1\right)$, since its expectation is 0, we can
still control its second moment using its gradient based on the
Poincar\'{e} inequality for marginal distribution.  The gradient of
this conditional expectation can be related to that of $h$, but there
will also be additional terms.  Drawing on the boundedness of
$\nabla_{x_1} \log \pi (x_2|x_1)$ and transformations of the integral,
we can show that these terms can also be controlled by the second
moment of gradient; therefore, we reach the conclusion of
Lemma~\ref{lemma-poincare-decomposition}.

Although log-concavity does not hold in the bad direction $\theta_0$,
the density remains unimodal within each partition, and we can make
use of the associated quasi-concavity.  In general, it is possible for
quasi-concave densities to have poor isoperimetry in high
dimensions~\cite{chandrasekaran2009sampling}. However, in one
dimension, we can lower bound the Cheeger constant as follows:
\begin{lemma}
  \label{lemma-quasi-concave-1d}
Assume that $\pi$ is a quasi-concave probability density function in
one-dimensional interval $[0,A]$.  For any partition $[0,A]=S_1\cup
S_2\cup S_3$ with $S_i$ being mutually disjoint, we have
    \begin{align*}
      \pi(S_3) \geq \frac{ \mathrm{dist}(S_1, S_2)}{A}
      \min(\pi(S_1), \pi(S_2)),
    \end{align*}
    where $\mathrm{dist}(S_1,S_2) \mydefn \inf \limits _{x_1\in
      S_1,x_2\in S_2}\Vert x_1-x_2\Vert$.
\end{lemma}
\noindent See Appendix~\ref{subsec:proof:lemma-quasi-concave-1d} for
the proof of Lemma~\ref{lemma-quasi-concave-1d}.

Equipped with Lemmas~\ref{lemma-poincare-decomposition} and
Lemma~\ref{lemma-quasi-concave-1d}, we can study the isoperimetry of
population power posterior on each partition.  In the next section, we
verify the structural assumptions used in the lemmas for symmetric
Gaussian mixtures, and establish the Cheeger constant for a large
bounded set in each partition.


\subsection{Isoperimetric inequalities for the population power posterior}
\label{subsec:iso_pop}

We now use the geometric tools in previous subsection to establish
isoperimetry for the density $\pi_0$ in one partition.  Throughout
this section, we assume without loss of generality that $\theta_{0} =
a_{0} e_1$ for $a_{0} > 0$ and $e_{1} = (1, 0, \ldots, 0)$. We take a
large cylinder $[0, A] \times \mathcal{B}(0, M)$, where the marginal
Poincar\'{e} inequality on $\Omega_1 = [0, A]$ and the conditional one
for $\Omega_2 = \mathcal{B}(0, M)$ can be established respectively.
Within these bounded domains, we also have a bound on the size of
gradient.  By taking $A$ and $M$ sufficiently large, we can guarantee
that most of the mass of $\pi_0$ is concentrated within $[- A, A]
\times \mathcal{B}(0, M)$.

\begin{lemma}
\label{corollary-structure}
For random vector $Y \sim \pi_0\propto e^{-U_0}$, we have
\begin{itemize}
\item The density of $Y_1$ exists and is quasi-concave
  on $\Rspace_+$.
\item For any $x_1 \in \Rspace_+$, the conditional density $Y_{- 1}| Y_1 = x_{1}$ 
  is log-concave on $\Rspace^{d - 1}$.
\end{itemize}
\end{lemma}
\noindent See Section~\ref{subsec:proof:corollary-structure}  for the
proof of Lemma~\ref{corollary-structure}. \\

A few comments are in order. First, Lemma~\ref{corollary-structure}
provides Cheeger constants for the conditional and marginal densities,
respectively.  In particular, by Lemma~\ref{lemma-quasi-concave-1d}, the
marginal density in the direction of the first standard basis vector
$e_1 \in \real^d$ has Cheeger constant lower bounded by $\frac{1}{A}$.
Additionally, by standard geometric results for log-concave densities
(cf. Theorem 2.4 in~\cite{lovasz2003logconcave}), the Cheeger
constant for the conditional density is lower bounded by $\frac{c}{M}$
for some universal constant $c > 0$.  Using Cheeger's inequality, these Cheeger constants
imply bounds for the Poincar\'{e} constants for the marginal density
and conditional density of $Y$ in Lemma~\ref{corollary-structure}.
Finally, invoking Lemma~\ref{lemma-poincare-decomposition} and these
Poincar\'{e} constants, we obtain the Poincar\'{e} constant for
$\pi_0$ on $[0, A] \times \mathcal{B}(0, M)$.  That Poincar\'{e}
constant also can be translated back to Cheeger constant via the
general results under negative curvature by Buser~\cite{buser1982note}
and Ledoux~\cite{ledoux1994simple}.  In light of these argument, the
conclusion of Theorem~\ref{theorem-isoperimetry-partition} is followed
(see Section~\ref{subsec:proof:theorem-isoperimetry-partition} for
proof details).
 
There is one caveat with our techniques. In fact, we require uniform
upper bounds of certain quantities in our analysis.  However, it does
not hold globally and can be shown to be satisfied in a reasonably
large region.  In order to address this technical challenge, we
truncate the tails of $\pi_{0}$ and $\pi$ as the probability outside
the large ball of these densities is exponentially decreasing.  The
truncation step necessitates use of the $s$-conductance framework for
Markov chains~\cite{lovasz2003logconcave} in order to translate from
isoperimetric results to conductance. These steps require tight tail
estimates for the densities $\pi_{0}$ and $\pi$.  In stating this
result, we introduce the quantity
\begin{align*}
  R_\varepsilon \mydefn C \left( 1 + \Vert \theta_0 \Vert +
  \frac{1}{\beta} \sqrt{d+ \log( 1/\varepsilon)} \right),
\end{align*}
and recall that $\pi_0(x) \propto e^{- U_0(x)}$ and $\pi(x) \propto
e^{-U(x)}$ denote the population and empirical power posteriors,
respectively.
\begin{proposition}
\label{corollary-large-ball}
There exists a constant $C > 0$ such that for any $\varepsilon \in
(0,1)$, the population power posterior $\pi_0$ satisfies
\begin{subequations}
  \begin{align}
 \pi_0 \left(\mathcal{B} \left(0, R_\varepsilon\right)\right) \geq 1 -
 \varepsilon.
    \end{align}
  Moreover, for any $\delta \in (0,1)$, given a sample size $\obs \geq
  (d + \Vert \theta_0 \Vert^2) \log \frac{ d + \Vert \theta_0 \Vert^2
  }{\delta}$, the empirical power posterior $\pi$ satisfies
  \begin{align}
      \pi \left(\mathcal{B} \left(0, R_\varepsilon\right)\right) \geq
      1 - \varepsilon.
  \end{align}
  with probability at least $1 - \delta$.
\end{subequations}
\end{proposition}
\noindent See Section~\ref{subsec:proof:corollary-large-ball} for the
proof of Proposition~\ref{corollary-large-ball}.

For any $s \in (0,1)$, by defining
\begin{align*}
A_s = M_s = R_{s} = C \left( 1 + \Vert \theta_0 \Vert + \frac{\sqrt{d+
    \log (1/ s)}}{ \beta } \right),
\end{align*}
we can guarantee that both $\pi$ and $\pi_0$ assign at least mass $1-
s$ to the Cartesian product \mbox{$[- A_{s}, A_{s}] \times
  \mathcal{B}(0, M_{s})$.}


\section{Bounds on the RMRW conductance}
\label{sec:conductance}

In this section, we analyze the $s$-conductance for the Markov chain
defined by RMRW algorithm (Algorithm~\ref{alg-MCMC-sym}), under both
$\pi_0$ and $\pi$, by using the geometric results in the previous
sections. By referring to the well-known connection between
conductance of Markov chain and its mixing time, we obtain the
conclusion of Theorem~\ref{theorem:mixing_time}. Furthermore, we also
establish the robustness of RMRW algorithm in terms of polynomial
mixing rate under model mis-specifcation.


\subsection{Markov chain conductance with population power posterior}
\label{subsection:population-conductance}

We first study the conductance of Markov chain defined by RMRW
algorithm when applied to the population quantities $U_0$ and $\pi_0$.
We denote $\mathcal{T}_{x, F}$ the transition kernel of that Markov
chain after Metropolis-Hasting step for any $F$. The following result
establishes a lower bound on the $s$-conductance of this Markov chain.
It involves the quantities
\begin{align*}
  A_s \defn M_s = C \left( 1 + \Vert \theta_0 \Vert + \frac{\sqrt{d+
      \log(1/ s)}}{\beta} \right), \quad \mbox{and} \quad \eta_s =
  \frac{1}{400 (A_s + M_s + \sqrt{d})^2},
\end{align*}
as well as the Cheeger constant bound $\cheeger_s = \frac{1}{\sqrt{d}
  A_{s}^5 M_{s}^2}$ from Theorem~\ref{theorem-isoperimetry-partition}.

\begin{proposition}
\label{cor-population-conductance}
There exists a universal constant $C > 0$ such that, for any set $S
\subseteq \Rspace^d$ such that $\pi_0( S) \in (s, \frac{1}{2})$, we
have
    \begin{align*}
        \int_{S} \mathcal{T}_{x, U_0}( S^{c}) \pi_0(x) dx \geq C
        \cdot( \pi_0( S) -s ) \sqrt{ \eta_{s} \cheeger_{s}}.
    \end{align*}
\end{proposition}
\noindent
See Section~\ref{subsec:proof:cor-population-conductance} for the
proof of Proposition~\ref{cor-population-conductance}.

Based on the
result of Proposition~\ref{cor-population-conductance}, the
$s$-conductance of Markov chain defined by RMRW algorithm with
$\pi_{0}$ and $U_{0}$ is lower bounded by $C \sqrt{\eta_s\cheeger_s}$
for any $s > 0$ where $\eta_{s}$ and $\cheeger_{s}$ are given in
Proposition~\ref{cor-population-conductance}.


\subsection{From population to sample power posterior}
\label{sec:sample}

As discussed in Section~\ref{sec:geometric-intuition}, the actual
power posterior for empirical data does not necessarily satisfy the
nice geometric properties of population power posterior.  In
particular, the mass within two basins-of-attraction can be
unbalanced.  In order to account for this problem, we establish a
uniform control for the difference between sample and population
posterior.  We denote $g_\theta(x) := \log f_{\theta} (x)$, $P_n
g_{\theta} \mydefn \frac{1}{n} \sum_{i = 1}^{n} g_{\theta}(X_{i})$,
and $P g_{\theta} \mydefn \Exs g_{\theta}(X)$ where the expectation is
taken with respect to symmetric Gaussian
mixtures~\eqref{eq:symmetric_Gaussian}.
\begin{lemma}
\label{lemma-empirical-process}
There exists a universal constant $c > 0$ such that, for any fixed constants $A , M \geq \Vert \theta_0 \Vert + 1$,  with
probability at least $1-\delta$ we have
\begin{align*}
  \sup_{\theta \in [- A, A] \times \mathcal{B}(0, M)}
  \left|P_n g_{\theta} - P g_{\theta} \right|
  \leq c \cdot (1 + A + M) \sqrt{ \frac{d}{n}
    \log \frac{n ( A + M + d)}{\delta}}.
\end{align*}
\end{lemma}
\noindent
The proof of Lemma~\ref{lemma-empirical-process} is in Section~\ref{subsec:proof:sec:proof-empirical-process}.  

Note that we
need $\bigO( d^2)$ sample complexity because the uniform concentration
is required in a global domain instead of local domain around
$\theta_{0}$. As seen from Proposition~\ref{corollary-large-ball}, the larger $\beta$ becomes, the smaller value of $A$ and $M$ we will have, and the mass will be more concentrated within a small region. Smaller region will make the uniform concentration bound tighter.

Equipped with Lemma~\ref{lemma-empirical-process}, we are able to derive the $s$-conductance of the Markov chain running with $U$ from its population counterpart, and prove the main theorem. By setting the sample size $n$ large enough, we can
guarantee that
\begin{align*}
   c^{-1} \pi_0( \theta) \leq \pi( \theta) \leq c \pi_0( \theta),
   \quad \text{and} \quad c^{ - 1} \mathcal{T}_{\theta, U_0}(S) \leq \mathcal{T}_{x,
     U}(S) \leq c \mathcal{T}_{\theta, U_0}(S),
\end{align*}
for some universal constant $c > 1$ and for any $\theta$ within the
cylinder $[- A, A] \times \mathcal{B}(0,M)$ such that $\theta \notin
S$.  These inequalities can be combined with
Proposition~\ref{cor-population-conductance} to obtain the
$s$-conductance of Markov chain defined by RMRW algorithm under $\pi$
and $U$.  Based on that result, we obtain the conclusion of
Theorem~\ref{theorem:mixing_time} for the polynomial mixing time of
RMRW algorithm (see Section~\ref{sec:proof-final} for detailed proof
of that theorem).



\subsection{Robustness under model mis-specification}

Recall that one of the main motivations for power posterior
distribution is its robustness under model
mis-specification~\cite{Miller-2019}. Accordingly, it is important to
show that the RMRW algorithm is robust with respect to model
mis-specification as well.  In order to analyze the performance of
RMRW algorithm in this setting, we use the following assumption:
\begin{assumption}
\label{assume-robustness}
Suppose that $\{X_i\}_{i=1}^n$ are i.i.d. samples from a contaminated
model with distribution $Q$, of the form $Q = (1 - \gamma) P_{0} +
\gamma F$, where $P_{0}$ is a probability distribution with density
$f_{\theta_{0}}$ and $F$ is a $K$-sub-Gaussian noise distribution for
some $K > 0$.
\end{assumption}
The contamination model specified in
Assumption~\ref{assume-robustness} is very flexible: the form of noise
distribution $F$ can be quite arbitrary, with only tail assumptions
needed.  This is actually very mild requirement, and even if it is not
satisfied, we can simply truncate the data points and enforce the
tail, without affecting the data from true distribution. The mixing
time of Algorithm~\ref{alg-MCMC-sym} appears to be robust with respect
to the model missepcification, as stated in the following proposition.
\begin{proposition}\label{prop-robust-misspecify}
   Assume that misspecified model satisfies
   Assumption~\ref{assume-robustness} and all other assumptions in
   Theorem~\ref{theorem:mixing_time} hold.  Then, with probability $1
   - \delta$, the mixing time bound from Theorem~\ref{theorem:mixing_time} still holds as
   long as $\gamma \leq \frac{c}{\beta(K^2 + 1) (d + \Vert \theta_0
     \Vert^2) \log n/\delta}$ for some universal constant $c>0$.
\end{proposition}
\noindent
The proof of Proposition~\ref{prop-robust-misspecify} is in Appendix~\ref{subsec:proof:prop-robust-misspecify}. 

A few comments are in order.  First, the basic idea of the proof is
straightforward: we relate the empirical power posterior $\pi \sim
e^{- U}$ defined by data from $Q$, to the "population" power posterior
defined by $P_{0}$, and the difference can be controlled with
$\gamma$. Note that, the population power posterior defined by $P_{0}$
is not the exact population power posterior distribution in the
contaminated model; however, since $\gamma$ is sufficiently small,
that population power posterior is very close to the true population
power posterior and is sufficient for the proof of
Proposition~\ref{prop-robust-misspecify}. Second, the result of
Proposition~\ref{prop-robust-misspecify} suggests that if the weight
associated with the contaminated distribution $F$ is $\bigO
(\frac{1}{\beta d})$, the RMRW algorithm is still able to achieve
polynomial mixing time for its convergence to the stationary power
posterior distribution.  Therefore, smaller value of $\beta$ not only
makes the power posterior itself more robust, but it also improves the
robustness of our RMRW algorithm.


\section{Proofs}
\label{sec:proofs}
In this section, we provide proofs for the key results with the
isoperimetry of population power posterior and mixing rate of Markov
chain in the paper.  In particular,
Section~\ref{subsec:isoperimetry_population_power} is devoted to
the proofs of results related to the isoperimetric inequalities that are given
in Section~\ref{sec:geometry}.  In Section~\ref{subsec:mix_rate_RMRW},
we provide the proofs for several results related to the mixing time
of Markov chain defined by RMRW algorithm in
Section~\ref{sec:conductance}.
\subsection{Proofs for isoperimetry of population power posterior}
\label{subsec:isoperimetry_population_power}
In this section, we provide proofs for results establishing
isoperimetric inequality for population power posterior of symmetric
Gaussian mixtures~\eqref{eq:symmetric_Gaussian}.

\subsubsection{Proof of Lemma~\ref{lemma-poincare-decomposition}}
\label{subsec:proof:lemma-poincare-decomposition}

For any function $h: \Omega_1 \times \Omega_2 \rightarrow \Rspace$
with $\int h(x)d\pi(x)=0$, we have
\begin{align}
  \int h^2(x)d\pi(x) & = \int_{\Omega_1} \int_{\Omega_2}
  h^2(x_1, x_2) d\pi_2 \big|_{x_1}(x_2) d\pi_1(x_1)
  \nonumber \\ & = \int_{\Omega_1}
  \left(\Exs_{\pi} \left(h(X_1, X_2)\Big|X_1 =
  x_1\right) \right)^2d \pi_1(x_1) \nonumber
  \\ & \hspace{12 em} + \int_{\Omega_1}
  \mathrm{var}_{\pi} \left(h(X_1, X_2) \Big|X_1 =
  x_1\right) d\pi_1(x_1). \label{eq:first_equ}
\end{align}
For the second term in equation~\eqref{eq:first_equ}, drawing on the
Poincar\'{e} inequality for the conditional distribution, we find that
\begin{align*}
  \mathrm{var}_{\pi} \left( h(X_1,X_2) \Big| X_1=x_1 \right) 
  \leq C_2 \int_{\Omega_2} \Vert \nabla_{x_2} h(x_1, x_2) \Vert^2 d\pi_2 \big|_{x_1}( x_2).
\end{align*}
Taking an integration of both sides of the above inequality leads to
\begin{align}
  \int_{ \Omega_1} \mathrm{var}_{\pi} \left(h(X_1,X_2) \Big|X_1=x_1\right) d\pi_1(x_1)
  \leq C_2 \int_{\Omega_1 \times \Omega_2} \Vert \nabla_{x_2} h(x_1,x_2) \Vert^2 d\pi(x_1,x_2). \label{eq:poincare_inequ_first}
\end{align}
For the first term in equation~\eqref{eq:first_equ}, note that we have
$\Exs \Exs_{\pi} \left(h(X_1, X_2) \big|X_1 \right) = 0$.  Invoking
the Poincar\'{e} inequality for the marginal distribution, we find
that
\begin{align}
  \int_{ \Omega_1} \left( \Exs_{ \pi} \left(h(X_1, X_2) \Big|X_1 = x_1
  \right) \right)^2d \pi_1(x_1) \leq C_1 \int_{\Omega_1} \left \Vert
  \nabla_{x_1} \Exs_{\pi} \left( h(X_1,X_2) \Big| X_1 = x_1 \right)
  \right \Vert^2 d\pi_1(x_1). \label{eq:poincare_inequ_second}
\end{align}
By the third condition in this lemma, the conditional distribution of
$X_2 | X_1$ has a density $\pi (x_2 | x_1)$ with respect to the
Lebesgue measure, and the function $x_1 \mapsto \log \pi (x_2 | x_1)$
is uniformly differentiable. By exchanging the derivative and
integration, the following equation holds
\begin{align*}
  \underbrace{\nabla_{x_1} \left( \int_{\Omega_2} h(x_1, x_2) \pi(x_2|x_1) dx_2 \right)}_{T_{1}}
  = \int_{\Omega_2} \nabla_{x_1} h(x_1, x_2) \pi(x_2| x_1) dx_2 & \\
  & \hspace{-8 em} + \underbrace{\int_{\Omega_2} h(x_1, x_2) \pi(x_2| x_1) \nabla_{x_1} \log \pi(x_2| x_1) dx_2}_{T_{2}}.
\end{align*}
By simple algebra, we arrive at the following identity
\begin{align*}
  T_{2} & = \int_{\Omega_2} \left(h(x_1, x_2) - \Exs_{\pi}
  \left(h(X_1, X_2) \Big| X_1 = x_1 \right) \right)
  \pi(x_2| x_1) \nabla_{x_1} \log \pi(x_2| x_1) dx_2 \\
  & \hspace{15 em} + \Exs_{\pi} \left(h(X_1, X_2) \Big|
  X_1 = x_1 \right) \nabla_{x_1} \left( \int_{ \Omega_2}
  \pi(x_2| x_1) dx_2 \right).
\end{align*}
Collecting the previous equations, we finally have
\begin{align*}
  T_{1} = \int_{\Omega_2} \left(h(x_1, x_2) - 
  \Exs_{\pi} \left(h(X_1, X_2) \Big|X_1 = x_1 \right)
  \right) \pi(x_2| x_1) \nabla_{x_1} \log \pi(x_2| x_1) dx_2 & \\
  & \hspace{-7 em} + \int_{\Omega_2} \nabla_{x_1} h(x_1, x_2)
  \pi(x_2| x_1) dx_2.
\end{align*}
Given the above equality, the following inequalities hold
\begin{align}
  & \int_{\Omega_1} \left \Vert \nabla_{x_1} \Exs_{\pi} \left(
  h(X_1,X_2) \Big| X_1 = x_1 \right) \right \Vert^2 d\pi_1(x_1)
  \nonumber \\ & \stackrel{(i)}{ \leq} 2 \int_{\Omega_1} \left \Vert
  \int_{\Omega_2} \nabla_{x_1} h(x_1, x_2) \pi(x_2|x_1) dx_2 \right
  \Vert^2 d\pi_1( x_1) \nonumber \\ & + 2 \int_{\Omega_1} \left \Vert
  \int_{ \Omega_2} \left(h(x_1, x_2) - \Exs_{\pi} \left(h(X_1,
  X_2) \Big|X_1 = x_1 \right) \right) \pi(x_2| x_1) \nabla_{x_1} \log
  \pi(x_2| x_1) dx_2 \right \Vert^2 d\pi_1(x_1) \nonumber \\ & 
  \stackrel{(ii)}{\leq} 2
  \int_{\Omega_1 \times \Omega_2} \Vert \nabla_{x_1} h(x_1, x_2)
  \Vert^2 d\pi(x_1, x_2) + 2 L^2
  \int_{\Omega_1} \mathrm{var}_{\pi} \left(h(X_1, X_2) \Big|X_1 = x_1
  \right) d\pi_1(x_1) \nonumber \\ & \stackrel{(iii)}{\leq} 2 \int_{
    \Omega_1 \times \Omega_2} \Vert \nabla_{x_1} h(x_1, x_2) \Vert^2
  d\pi(x_1, x_2) + 2 C_2 L^2
  \int_{\Omega_1 \times \Omega_2} \Vert \nabla_{x_2} h(x_1, x_2)
  \Vert^2 d\pi(x_1, x_2), \label{eq:key_ineq}
\end{align}
where inequality (i) follows from Young's inequality; inequality in
(ii) follows from the uniform upper bound of $\vecnorm{\nabla_{x_1}
  \log \pi (x_2| x_1)}{2}$, and the inequality (iii) is based on
Poincar\'{e} inequality for the conditional distribution. Putting
together the results of equations~\eqref{eq:first_equ},
~\eqref{eq:poincare_inequ_first}, ~\eqref{eq:poincare_inequ_second},
and~\eqref{eq:key_ineq}, we obtain that
\begin{align*}
  \int h^2(x) d\pi(x) & \leq C_2 \int \Vert \nabla_{x_2}h(x_1, x_2)
  \Vert^2 d\pi(x_{1}, x_{2}) + 2 C_1 \int \Vert \nabla_{x_1}h(x_1,
  x_2) \Vert^2 d\pi(x_{1}, x_{2})\\ &\quad \quad \quad+ 2C_1 C_2 L^2
  \int \Vert \nabla_{x_2} h(x_1, x_2) \Vert^2 d\pi(x_{1}, x_{2})\\ &
  \leq 2 \parenth{C_1 + C_2 + C_1 C_2 L^2}\int \Vert\nabla h(x)\Vert^2
  d\pi(x),
\end{align*}
which completes the proof of the lemma.

\subsubsection{Proof of Lemma~\ref{corollary-structure}}
\label{subsec:proof:corollary-structure}
Before the proof, we introduce and recall a few notation that 
we will use throughout this section. 
For a vector $z \neq 0$, let $z^{\perp}$ denote the subspace orthogonal to $z$, 
i.e., $z^\perp \mydefn \{x: z^{\top} x = 0\}$. 
For any vector $x \in \real^d$, let $x_i$ denote the $i$-th coordinate of $x$, 
and $x_{-i}$ denote the $(d-1)$-dimensional vector $(x_1, \ldots, x_{i-1}, x_{i+1}, \ldots, x_d)$. 
For a $d$-dimensional density $\pi$, we use $\pi^{(i)}$
to denote its marginal density on the $i$-th coordinate, 
and $\pi^{(-i| i)}$ to denote the conditional density of $X_{-i}$ 
conditioned on $X_i$ with $X \sim \pi$.

 We study the geometry of logarithm of the population power posterior by direct calculation. 
 For $X \sim \frac{1}{2} \mathcal{N} (\theta_{0}, I_{d}) 
 + \frac{1}{2} \mathcal{N} (- \theta_{0}, I_{d})$, 
 we can write $X$ as $X = a_{0} \tau e_1 + \xi$ 
 with $\xi \sim \mathcal{N} (0, I_{d})$ and $\tau$ being a Rademacher random variable independent of $\xi$.
 Straightforward calculation leads to the following results
\begin{align} \label{eq-population-derivatives}
    \nabla U_0( \theta)
    = & \beta \theta + \beta\Exs \left(\frac{- \varphi(X - \theta) 
    + \varphi(X + \theta)}{\varphi(X - \theta) +
    \varphi(X + \theta)} X\right) \nonumber \\
    = & \beta \theta + \frac{\beta}{2}
    \Exs \left( \frac{- \varphi(a_0 e_1 + \xi - \theta)
    + \varphi(a_0 e_1 + \xi + \theta)}{\varphi(a_0 e_1 + 
    \xi - \theta) + \varphi(a_0 e_1 + \xi + \theta)}
    (a_0 e_1 + \xi) \right) \nonumber \\
    + & \frac{\beta}{2} \Exs \left( \frac{ - 
    \varphi(-a_0 e_1 + \xi - \theta) + \varphi(- a_0 e_1 + \xi + \theta)}
    {\varphi(- a_0 e_1 + \xi - \theta) + \varphi(-a_0 e_1 + \xi + \theta)}
    (- a_0 e_1 + \xi) \right) \nonumber \\
    = & \beta \theta + \beta \Exs \left( \frac{ - 
    \varphi(a_0 e_1 + \xi - \theta) + \varphi(a_0 e_1 + \xi + \theta)}
    {\varphi(a_0 e_1 + \xi - \theta) + \varphi(a_0 e_1 + \xi + \theta)}
    (a_0 e_1 + \xi) \right), \quad \text{and} \nonumber \\ 
    \nabla^2 U_0(\theta)
    = & \beta I_{d} - 4 \beta \Exs \left ( \frac{ 
    \varphi(X - \theta) \varphi (X + \theta)} {( \varphi(X - \theta) 
    + \varphi(X + \theta))^2} X X^{\top} \right) \nonumber \\
    = & \beta I_{d} - a_0^2 \beta \Exs \left( \frac{4 \varphi(X - \theta)
    \varphi(X + \theta)}{(\varphi(X - \theta) + \varphi(X
    + \theta))^2} \right) e_1 e_1^{\top} \nonumber \\
    - & 
    \beta \Exs \left( \frac{4 \varphi(X - \theta) 
    \varphi(X + \theta)}{(\varphi(X - \theta) + \varphi(X
    + \theta))^2} \xi \xi^{\top} \right).
\end{align}
In order to obtain the conclusion of the proposition, we need the following lemma:
\begin{lemma}\label{lemma-convex-quasi-concave}
Let subspace $H \mydefn \{e_1\}^\perp$. For the geometry of $U_0$, we have
\begin{itemize}
    \item The function $f_1: \real_+\rightarrow \real$ defined by $f_1(a)=U_0( ae_1 + z)$ is non-decreasing on $a \in [0, a_0]$ and non-increasing on $a \in [a_0, +\infty)$, for any $z\in H$.
    \item The function $f_2 : H \rightarrow \real$ defined by $f_2(z)=U_0( ae_1 + z)$ is a convex function of $z$, for any $a>0$.
\end{itemize}
\end{lemma}
\noindent
The proof of Lemma~\ref{lemma-convex-quasi-concave} is in Appendix~\ref{subsec:proof:lemma-convex-quasi-concave}. 

Equipped with that lemma, we are ready to prove the structural results about geometry of $\pi$. First, for the marginal density along the direction of $\theta_0$, 
we note that $u \mapsto e^{- u}$ is a monotonic decreasing function on $\real$. 
By Lemma~\ref{lemma-convex-quasi-concave}, we can conclude that for any 
$z \in e_1^\perp$, $\pi(a e_1 + z) \propto e^{- U_0(a e_1 + z)}$ 
is an increasing function on $[0, a_0]$ and a decreasing function on $[a_0, +\infty)$. Thus, we obtain that
 \begin{align*}
        \pi_0^{ (1)}(a) 
        = \int_{ e_1^{ \perp}} \pi(a e_1 + z) dz,
    \end{align*}
which is an average of increasing functions on $[0, a_0]$,
and an average of decreasing functions on $[a_0, +\infty)$. 
Therefore, it is also increasing on $[0, a_0]$ and decreasing on $[a_0, +\infty)$, 
which means it is quasi-concave. 

For the conditional density of $X_{- 1}| X_{1}$, we have the following equation
\begin{align*}
  \log \pi_0^{(-1| 1)}(x_{ - 1}| x_1)
  = \log \pi_0([0, x_{- 1}] + x_1 e_1) - \log \int_{e_1^{\perp}} \pi_0(x_1 e_1 + z) dz.
\end{align*}
It is clear that the second term in the above display is independent
of $x_{-1}$.  Therefore, the conditional density of $X_{- 1}| X_{1}$
is log-concave on $\Rspace^{d - 1}$.  As a consequence, we have
established the claim in the proposition.


\subsubsection{Proof of Theorem~\ref{theorem-isoperimetry-partition}}
\label{subsec:proof:theorem-isoperimetry-partition}

In this section, we prove Theorem~\ref{theorem-isoperimetry-partition}
by combining the structural results in the paper together.  First,
note that Lemma~\ref{lemma-poincare-decomposition} provides a way of
combining Poincar\'{e} constant estimates for the marginal and
conditional distribution, and Lemma~\ref{lemma-quasi-concave-1d} gives
lower bound on Cheeger constant of the marginal distribution. For the
conductance framework in Markov chain mixing results, we also need
isoperimetry in the form of Cheeger constants.  Therefore, we utilize
existing results that relate Cheeger constants and Poincar\'{e}
constants in both directions.

Translating from the Cheeger inequality to the Poincair\'{e}
inequality is relatively easy due to the well-known Cheeger
inequality~\cite{cheeger1969lower}: the Poincar\'{e} constant can be
controlled using the Cheeger constant: $\poincare_\pi \leq \frac{4}{
  \cheeger_\pi^2}$ for any density $\pi$ on $\real^d$.

The argument in the other direction is more involved.  We make use of
the result in the work of Buser~\cite{buser1982note} and
Ledoux~\cite{ledoux1994simple}. The result was first proven for
manifolds with a uniform Ricci curvature lower bound. Ledoux's proof
is based upon the Li-Yau inequality, which also works for
Bakry-Emery-Ricci curvature of diffusion semigroups.
\begin{proposition}
  \label{prop-buser}
  Consider a $d$-dimensional Riemannian manifold $\mathcal{M}$,
  equipped with density $e^{- \Psi}$.  Let $\lambda_1$ be the eigengap
  of the semigroup generator $\Delta + \nabla \Psi \cdot \nabla$ and
  $\cheeger$ be the Cheeger constant of the measure $e^{- \Psi}$ on
  $\mathcal{M}$. Suppose the Bakry-Emery-Ricci Curvature is uniformly
  lower bounded, meaning that $\mathrm{Ric} + \nabla^2 \Psi \succeq
  -K$ for some constant $K > 0$. Then we have
  \begin{align*}
    \lambda_1\leq 10 \sqrt{d} (\cheeger \sqrt{K} + \cheeger^2).
  \end{align*}
\end{proposition}

With these technical tools at our disposal, we are ready to establish
the geometric result.  The proof of the theorem follows from the
applications of several results in the paper.  In particular, we first
combine Lemma~\ref{corollary-structure} with
Lemma~\ref{lemma-quasi-concave-1d} and localization-based isoperimetry
tools~\cite{bobkov1999isoperimetric} to derive Poincar\'{e}
inequalities for the marginal and conditional densities. These
inequalities are combined together to obtain the Poincar\'{e}
inequality for $\pi_0$ on this region.  Then, we can apply
Proposition~\ref{prop-buser} to get the Cheeger constant.

In order to apply Proposition~\ref{prop-buser}, we need to lower bound
the negative curvature of $U_{0}( \theta)$. For any $\theta
\in \Rspace^d$, from equation~\eqref{eq-population-derivatives}, we
find that
\begin{align*}
  \nabla^2 U_0( \theta) = & \beta I_{d} - a_0^2 \beta \Exs
  \parenth{\frac{4 \varphi(X - \theta) \varphi(X + \theta)}{(\varphi(X
      - \theta) + \varphi(X + \theta))^2}} e_1 e_1^{\top} - \beta \Exs
  \parenth{\frac{4 \varphi(X - \theta) \varphi( X + \theta)}
    {(\varphi(X - \theta) + \varphi(X + \theta))^2} \xi \xi^{\top}}
  \\ \succeq & \beta I_{d} - a_0^2 \beta \Exs \parenth{\frac{4
      \varphi(X - \theta) \varphi(X + \theta)} {(\varphi(X - \theta) +
      \varphi(X + \theta))^2}} e_1 e_1^{\top} - \beta \Exs
  \parenth{\xi \xi^{\top}} \\ \succeq & -a_0^2 \beta I_{d}.
\end{align*}
Therefore, the negative curvature of $U_{0}$ is bounded from below by
$ - \beta \Vert \theta_0 \Vert^2$.

Lemma~\ref{lemma-poincare-decomposition} requires a uniform upper
bound on the gradient of conditional density, which holds true in a
bounded region:
\begin{align*}
  \abss{\frac{\partial}{\partial \theta_1} \log \pi_0^{(-1|1)}
    (\theta^{(- 1)}| \theta^{(1)})} \leq & \abss{
    \frac{\partial}{\partial \theta_1}\log\pi_0(\theta)} + \abss{
    \frac{ \partial}{ \partial \theta_1} \log \int_{e_1^\perp} \pi_0
    (z + \theta^{(1)} e_1) dz} \\ \leq & 2 (|\theta^{(1)}| +
  \Exs(|a_0| + |\xi_1|)) \\ \leq & 2 A + 2 \Vert \theta_0 \Vert + 4.
\end{align*}
By Lemmas~\ref{lemma-quasi-concave-1d} and~\ref{corollary-structure},
the Cheeger constant for the marginal density of $\pi_0$ on the
direction $e_1$ is lower bounded by $1/ A$.  Furthermore, by
Lemma~\ref{corollary-structure} and classical results for log-concave
densities~\cite{bobkov1999isoperimetric}, the Cheeger constant for
the conditional density on other directions is lower bounded by
$\frac{1}{2 M}$.  Invoking the Cheeger inequality leads to
Poincar\'{e} constants $4A^2$ and $16M^2$ respectively.
	
By Lemma~\ref{lemma-poincare-decomposition}, the Poincar\'{e} constant
for $\pi_0$ is upper bounded by:
\begin{align*}
  \poincare \leq 2(4 A^2 + 16 M^2 + 64 A^2 M^2(2 A + 2 \Vert \theta_0
  \Vert + 4)^2) \lesssim A^4 M^2.
\end{align*}
Using Proposition~\ref{prop-buser} with $K= \beta \Vert
\theta_0\Vert^2$, we obtain that $\cheeger \gtrsim \frac{1}{\sqrt{d}
  A^5 M^2}$. As a consequence, we reach the conclusion of the theorem.


\subsubsection{Proof of Proposition~\ref{corollary-large-ball}}
\label{subsec:proof:corollary-large-ball}

In order to establish the tail bounds for $\pi_{0}$ and $\pi$, we
first prove a more general result that applies to a broader class, one
which includes $\pi_0$ and $\pi$.
\begin{lemma}
  \label{lemma-dissipative-tail}
Consider a differentiable function $\bar{ U}$ such that
$\inprod{x}{\nabla \bar{ U}(x)} \geq a \Vert x \Vert^2 - b$ for any $x
\in \Rspace^d$. Then there exists a universal constant $C > 0$ such
that for any $\delta \in (0,1)$, we have
\begin{align*}
  \mprob_{\pi} \left[ \|X\| \geq C \sqrt{ \frac{ b + d + \log
        \frac{1}{ \delta}}{a}} \right] \leq \delta,
  \end{align*}
where $\mprob_\pi$ denotes the probability under the density function
$\pi(x) \propto e^{- \bar{ U}(x)}$.
\end{lemma}
\noindent
The proof of Lemma~\ref{lemma-dissipative-tail} is in
Appendix~\ref{subsec:proof:lemma-dissipative-tail}.

Now we can bound the tails of both $\pi_0$ and $\pi$ by validating the
conditions needed in Lemma~\ref{lemma-dissipative-tail}. In fact, for
$U_0$ we have
\begin{align*}
  \langle \nabla U_0(\theta), \theta \rangle = & \beta \|\theta \|^2 +
  \beta \Exs \left( \frac{ - \varphi(a_0 e_1 + \xi - \theta) +\varphi(
    a_0 e_1 + \xi + \theta)} {\varphi(a_0 e_1 + \xi - \theta) +
    \varphi(a_0 e_1 + \xi + \theta)} \theta^{\top} (a_0 e_1 + \xi)
  \right) \\
 \geq & \beta \|\theta\|^2 - a_0 \beta \| \theta\| - 2 \beta
 \Exs \left( \frac{ \varphi( a_0 e_1 + \xi - \theta)}
        {\varphi(a_0 e_1 + \xi - \theta) + \varphi(a_0 e_1 + \xi +
          \theta)} \theta^{\top} \xi \right).
\end{align*}
For a vector of the form $\theta = a e_1 + z$ with $z \perp e_1$, we
find that
\begin{align*}
  \Exs \left (\frac{ \varphi(a_0 e_1 + \xi - \theta)} {\varphi(a_0 e_1
    + \xi - \theta) + \varphi(a_0 e_1 + \xi + \theta)} \theta_{-
    1}^{\top} \xi_{-1} \right) & = \Exs \left(\frac{1}{1 + \exp
    \left(- 4 (a_0 + \xi_1) a - 4 z_{-1}^{\top} \xi_{- 1}\right)}
  z^{\top} \xi_{- 1} \right) \\ & \geq 0,
\end{align*}
which is due to the symmetry of the law of $z^{\top}
\xi_{-1}$. Therefore, the following inequalities hold
\begin{align*}
  \langle \nabla U_0( \theta), \theta \rangle & \geq \beta
  \Vert \theta \Vert^2 - a_{0} \beta \Vert \theta \Vert - 2 \beta
  \Exs| \xi_1| \cdot \Vert \theta \Vert \\ & \geq
  \frac{\beta}{2} \Vert \theta \Vert^2 - \beta(\Vert \theta_0
  \Vert^2 + 1).
\end{align*}
Hence, we achieve the conclusion of the proposition with $U_{0}$.

For $U$, we utilize empirical process tools to establish the
dissipativity condition. In particular, we denote $L(\theta; X_{i}) =
\frac{1}{2} \varphi( \theta - X_i) + \frac{1}{2} \varphi(\theta +
X_i)$ for all $i \in [n]$. Then, we have the following equations
    \begin{align*}
      \langle \nabla \log L(\theta; X_i), \theta \rangle & = \Vert
      \theta \Vert^2 - \frac{\varphi(X_i - \theta) - \varphi(X_i +
        \theta)}{\varphi(X_i - \theta) + \varphi(X_i + \theta)}
      \theta^{\top} X_i \\ & = \Vert \theta \Vert^2 - \frac{e^{ 
          \theta^{\top} X_i} - e^{- \theta^{\top} X_i}}{e^{
          \theta^{\top} X_i} + e^{- \theta^{\top} X_i}} \theta^{\top}
      X_i.
    \end{align*}
 Define $Z \mydefn \sup_{\Vert \theta \Vert \leq R} \frac{1}{n}
 \sum_{i = 1}^n \sigma_i \inprod{ \nabla \log L(\theta; X_i)}{ \theta}
 $. Note that the function $z \mapsto \frac{e^{ z} - e^{ - z}} {e^{ z
   } + e^{- z}} z$ is centered and Lipschitz with constant $L =
 3$. Therefore, by the Ledoux-Talagrand contraction inequality for
 Lipschitz functions of Rademacher
 processes~\cite{Ledoux_Talagrand_1991}, we have
\begin{align}
  \Exs \brackets{ Z} \leq 6 \Exs \brackets{ \sup_{\Vert \theta \Vert
      \leq R} \frac{1}{n} \sum_{i = 1}^n \sigma_i \langle X_i, \theta
    \rangle } \leq 6R\sqrt{\frac{ d + \Vert \theta_0 \Vert^2 }{ n
  }}. \label{eq:bound_Z}
\end{align}
In order to obtain a high-probability bound for $Z$, we apply a
functional Bernstein inequality due to a Talagrand, after a suitable
truncation-based argument.  Beginning with the truncation step, for
some $b > 0$ to be chosen, let us define the event
\begin{align*}
    \Event_b \mydefn \left\{ \max_{1 \leq i \leq n} \sup_{\Vert \theta
      \Vert \leq R} |g_\theta (X_i)| \leq b \right\}.
\end{align*}
We then apply Talagrand's theorem on empirical processes (Theorem 3.27
in the book~\cite{Wai19}), conditionally on this event, so as to
obtain
\begin{align}
  \label{EqnTalagrand}
      \Prob(Z > \Exs \brackets{ Z} + t \mid \Event_b) \leq \exp \left(
      - \frac{n t^2}{56 \Exs \brackets{ \Sigma^2} + 4 b t} \right),
  \end{align}
where $\Sigma^2 \mydefn \sup_{\Vert \theta \Vert \leq R} \frac{1}{n}
\sum_{i = 1}^n g_\theta^2 (X_i)$.  For any $n \geq d$, we have
\begin{align*}
    \Exs \brackets{ \Sigma^2} & \leq R^2 + \Exs \brackets{ \sup_{\Vert
        \theta \Vert \leq R} \frac{1}{n} \sum_{i = 1}^n (\theta^{
        \top} X_i \tanh (\theta^{ \top} X_i))^2} \\
& \leq R^2 + \Exs \brackets{ \sup_{\Vert \theta \Vert \leq R}
      \frac{1}{n} \sum_{i = 1}^n (\theta^{ \top} X_i )^2 } \\
    & = R^2 \left(1 + \Exs \opnorm{\frac{1}{n} \sum_{i = 1}^n X_i
      X_i^{ \top}} \right)\\
&    \leq R^2 (3 + \Vert \theta_0 \Vert^2).
\end{align*}

By standard $\chi^2$ concentration results (Example
2.11,~\cite{Wai19}), with probability at least $1 - \delta$, we have
\begin{align*}
    \Prob (\Vert X_i \Vert \geq 2 \Vert \theta_0 \Vert + 2 \sqrt{d} + 2t) \leq e^{- 2 t^2},
\end{align*}
for any $t > 0$ and $i \in [n]$. In addition, we find that
\begin{align*}
    |g_\theta (x)| \leq \Vert \theta \Vert^2 + \Vert \theta \Vert \cdot \Vert x \Vert 
    \leq R^2 + R \Vert x \Vert \quad \text{for all} \ x.
\end{align*}
Putting the above results together, by letting $b \mydefn 
R^2 + 2R ( \Vert \theta_0 \Vert +  \sqrt{d} + \sqrt{\log 2 n / \delta})$ we obtain that
\begin{align}
  \label{EqnEbound}
    \Prob(\Event_b^c) \leq n \cdot \Prob \left( R^2 + R \Vert X_1
    \Vert > b \right) \leq \delta/2.
\end{align}

Setting $t = 112 \sqrt{\frac{\Exs \brackets{ \Sigma^2} \log
    \delta^{-1}}{n}} + 8 \frac{b \log \delta^{-1}}{n}$ in our earlier
bound~\eqref{EqnTalagrand} and combining with the upper
bound~\eqref{EqnEbound} on $\Prob (\Event_b^c)$, we find that
  \begin{align*}
      \Prob \left( Z \geq \Exs \brackets{ Z} + c \cdot R \sqrt{\frac{
          (1 + \Vert \theta_0 \Vert^2) \log \delta^{-1}}{n}} + c \cdot
      \frac{R^2 + \Vert \theta_0 \Vert^2 + d + \log
        \frac{2n}{\delta}}{n} \log \delta^{-1} \right) \leq \delta,
  \end{align*}
  for some universal constant $c > 0$. Invoking a symmetrization
  inequality for probabilities~\cite{vanderVaart-Wellner-96}, we have
  \begin{align*}
      \Prob \left(\sup_{\Vert \theta \Vert \leq R}|(P - P_n) g_\theta|
      > t \right) \leq c_1 \Prob (Z > c_2 t), \quad \text{for all} \ t
      > 0,
  \end{align*}
  for some universal constants $c_1, c_2 > 0$.
  
  For any $R> 0, \delta > 0$, we define
  \begin{align*}
      \Delta_n (R, \delta) \mydefn R \sqrt{\frac{ d + (1 + \Vert
          \theta_0 \Vert^2) \log \delta^{-1}}{n} } + \frac{R^2 + \Vert
        \theta_0 \Vert^2 + d + \log \frac{2n}{\delta}}{n}\log
      \delta^{-1}.
  \end{align*}
  Combining the previous inequalities with the upper bound on $\Exs
  \brackets{ Z}$ in equation~\eqref{eq:bound_Z}, we have the following
  bound
  \begin{align}
      \Prob \left( \sup_{\Vert \theta \Vert \leq R}|(P - P_n)
      g_\theta| \geq c \cdot \Delta_n (R, \delta) \right) \leq
      \delta, \label{eq:tail-for-dissipative}
  \end{align}
  for a universal constant $c > 0$.

For $\delta > 0$ fixed, let $\delta_k = \frac{ 6 \delta }{ k^2 \pi^2
}$ with $\sum_{k = 1}^{\infty} \delta_k = \delta$, and let $R_k=2^k$.
Combining the union bound with the ensemble of
inequalities~\eqref{eq:tail-for-dissipative} for $R = R_k,
~k=1,2,\ldots$, we find that
\begin{multline*}
  \mprob \left( \exists \ \theta \in \real^d \mbox{ such that} \ |(P_n
  - P) g_\theta (X)| \geq c \cdot \Delta \left( 2 \Vert \theta \Vert,
  \frac{6 \delta}{\pi^2 \log^2 (1 + 2 \Vert \theta \Vert)} \right)
  \right)\\ \leq \sum_{k = 1}^{+\infty} \Prob \left( \sup_{\Vert
    \theta \Vert \leq R}|(P - P_n) g_\theta| \geq c \cdot \Delta_n
  (R_k, \delta_k) \right) \leq \delta,
\end{multline*}
where $c$ is some universal constant. Using standard $\chi^2$ tail
bounds, we find that
\begin{align*}
    \frac{1}{n}\sum_{i = 1}^n \Vert X_i \Vert^2 \leq \Vert \theta_0
    \Vert^2 + 2 \sqrt{\frac{ d +  \log \delta^{- 1}}{n}}
\end{align*}
with probability $1 - \delta$. Therefore, for $n \geq c' ( d + \Vert
\theta_0 \Vert^2 ) \log \frac{ d + \Vert \theta_0 \Vert^2}{\delta}$
for some universal constant $c' > 0$, with probability $1 - \delta$,
we obtain that
\begin{align*}
    \langle \nabla U(\theta), \theta \rangle 
    \geq \frac{\beta}{2} \Vert \theta \Vert^2 
    - 2\beta \left( \Vert\theta_0\Vert^2 + 1 \right).
\end{align*}
Plugging this result into Lemma~\ref{lemma-dissipative-tail} yields
the claim in the proposition.


\subsection{Proofs related to mixing rates}
\label{subsec:mix_rate_RMRW}

In this section, we provide the proofs for several results related to
the mixing time of Markov chain generated by the RMRW algorithm
(Algorithm~\ref{alg-MCMC-sym}).


\subsubsection{Proof of Proposition~\ref{cor-population-conductance}}
\label{subsec:proof:cor-population-conductance}

Let $\mathcal{P}_{x, F}$ denote the unadjusted proposal distribution
of the RMRW algorithm when at position $x$.  Recall that
$\mathcal{T}_{x, F}$ denotes the transition kernel of the Markov chain
defined by the RMRW algorithm after the Metropolis-Hasting correction.
The following lemma provides bounds needed for controlling the
$s$-conductance of the Markov chain at the level of the population-level
density $\pi_0$:
\begin{lemma}
  \label{lemma-rejection-overlap}
  Consider a scalar $\eta \in \big(0, \frac{1}{400 (A + M +
    \sqrt{d})^2} \big)$ and a pair $x, y \in [- A, A] \times
  \mathcal{B}(0, M)$ such that $\max( \Vert x - y \Vert, \Vert x + y
  \Vert) \leq \frac{1}{10} \sqrt{\eta}$.  We then have the bounds
 \begin{align*}
   \dtv( \mathcal{T}_{x, U_0}, \mathcal{P}_{x,U_0}) \leq \frac{1}{10},
   \quad \text{and} \quad \dtv( \mathcal{T}_{x, U_0}, \mathcal{T}_{y,
     U_0}) \leq \frac{1}{2}.
    \end{align*}
\end{lemma}
\noindent
See Appendix~\ref{subsec:lemma:lemma-rejection-overlap} for the proof
of this claim.

Using Lemma~\ref{lemma-rejection-overlap}, we can now complete the
proof of the proposition. Since $\pi_0$ and $T_{x, U_0}$ are both
symmetric around 0, we only need to control the $s$-conductance of the
Markov chain on each partition.  Let us introduce the truncated
quantities
    \begin{align*}
        \tilde{ \pi}_0(x) \mydefn & \begin{cases} 2 \pi_0(x) & x_1 > 0
          \\ 0 & \text{otherwise} \end{cases}, \quad \text{and}
        \\ \tilde{T}_{x,U_0}(y) \mydefn & \begin{cases} T_{x, U_0}(y)
          + T_{x, U_0}(- y) & y_1 > 0 \\ 0 &
          \text{otherwise} \end{cases}.
    \end{align*}
The claim of the proposition is equivalent to the inequalities for
$\tilde{\pi}_0$ and $\tilde{T}_{x, U_0}$ restricted to the cylinder
$[0, A] \times \mathcal{B}(0, M)$.  We combine
Lemma~\ref{lemma-rejection-overlap} with the same argument as that in
the proof of Lemma 2 from the paper~\cite{dwivedi2018log}.  In
particular, we replace the isoperimetric inequality in the latter
lemma with that from Theorem~\ref{theorem-isoperimetry-partition}.
Doing so yields the desired bound on the $s$-conductance of the Markov
chain defined by RMRW algorithm, thereby completing the proof of the
proposition.

\subsubsection{Proof of Lemma~\ref{lemma-empirical-process}}
\label{subsec:proof:sec:proof-empirical-process}

For any fixed $\theta \in [-A, A] \times \mathcal{B}(0, M)$, due to
the symmetry of $\xi \sim \mathcal{N}(0, I_{d})$, we have
    \begin{align*}
        \frac{1}{2} \varphi( \theta + \theta_0 + \xi) + 
        \frac{1}{2} \varphi(\theta - \theta_0 - \xi) 
        \overset{d}{=}
        \frac{1}{2} \varphi(\theta - \theta_0 + \xi) + 
        \frac{1}{2} \varphi(\theta + \theta_0 - \xi),
    \end{align*}
The above distribution equation leads to
    \begin{align}
        g_\theta( \xi)
        & = \log \left ( \frac{1}{2} \varphi( \theta + \tau
        \theta_0 + \xi) + \frac{1}{2} \varphi( \theta - \tau 
        \theta_0 - \xi) \right) \nonumber \\
        & \overset{d}{=} \log \left( \frac{1}{2} \varphi( \theta
        - \theta_0 + \xi) + \frac{1}{2} \varphi( \theta +
        \theta_0 - \xi) \right), \label{eq-equal-in-distr-emp}
    \end{align}
where $\tau$ is a Rademacher random variable independent of $\xi$. The
equal in distribution result in equation~\eqref{eq-equal-in-distr-emp}
is uniform in $\theta$. Namely, $g_\theta(\xi)$ as a random function
of $\theta$ has the same distribution as the term $\log \left(
\frac{1}{2} \varphi( \theta - \theta_0 + \xi) + \frac{1}{2} \varphi(
\theta + \theta_0 - \xi) \right)$. We denote
\begin{align*}
    h_{\theta_0, \theta}( \xi) \mydefn \log \left( \frac{1}{2}
    \varphi(\theta - \theta_0 + \xi) + \frac{1}{2} \varphi(\theta +
    \theta_0 - \xi) \right) + \frac{1}{2} \Vert \xi \Vert^2.
\end{align*} 
Following some simple algebra, we find that
    \begin{align*}
      \Vert \nabla_\xi h_{ \theta_0, \theta}( \xi) \Vert = \Vert
      \frac{( \theta_0 - \theta) \varphi( \theta - \theta_0 + \xi) +
        (\theta_0 + \theta) \varphi( \theta + \theta_0 - \xi)}
           {\varphi( \theta - \theta_0 + \xi) + \varphi( \theta +
             \theta_0 - \xi)} \Vert \leq \Vert \theta_0 \Vert + \Vert
           \theta \Vert \leq 2 (A + M).
    \end{align*}
Since $h_{\theta_0, \theta}(\xi)$ is a Lipschitz function of Gaussian
random vector, invoking a classical concentration inequality
(cf. Theorem 2.26,~\cite{Wai19}) yields
    \begin{align*}
       \mprob \left( \biggr|h_{\theta_0, \theta}( \xi) - \Exs h_{
         \theta_0, \theta}( \xi)\biggr| \geq t \right) \leq 2 \exp
       \left( - \frac{t^2}{8 (A + M)^2} \right)
    \end{align*}
for all $t > 0$.  In words, we have shown that the random variable
$h_{\theta_0, \theta}( \xi)$ is sub-Gaussian with parameter $4 ( A +
M)$.  For $\mathrm{i.i.d.}$ copies $\xi_{1}, \ldots, \xi_{n}$ from
$\NORMAL(0, I_{d})$, we find that
    \begin{align*}
         \mprob \left( \abss{ \frac{1}{n} \sum_{i = 1}^n
         h_{\theta_0, \theta}( \xi) - \Exs h_{\theta_0,
         \theta}( \xi)} \geq t \right) 
         \leq 2 \exp \left( - \frac{n t^2}{8 (A + M)^2} \right)
    \end{align*}
for all $t > 0$.

Now we turn to bound the uniform concentration of
$h_{\theta_0,\theta}$. For some $\varepsilon > 0$, whose specific
value will be determined later, let $\mathcal{M}_\varepsilon =
\{\theta_1, \theta_2, \cdots, \theta_{N_\varepsilon} \}$ be an
$\varepsilon$-covering of the cylinder $[-A, A] \times \mathcal{B}(0,
M)$.  Note that, for $\theta_1, \theta_2 \in [- A, A] \times
\mathcal{B}(0, M)$, we have
    \begin{align*}
        |P_n h_{\theta_0, \theta_1} - P_n h_{\theta_0, \theta_2}| \leq
        & \Vert \theta_1 - \theta_2 \Vert \parenth{ \frac{1}{n}
          \sum_{i = 1}^n \sup_{\theta \in [- A, A] \times
            \mathcal{B}(0, M)} \Vert \nabla_{\theta} g_\theta(X_i)
          \Vert} \\ \leq & \Vert\theta_1 - \theta_2 \Vert \cdot
        \left(A + M + \frac{1}{n} \sum_{i = 1}^n \Vert X_i \Vert
        \right).
    \end{align*}
For any $\theta\in[- A, A] \times \mathcal{B}(0, M)$, denote $\xi(
\theta) \mydefn \arg \min_{\theta' \in \mathcal{M}_{ \varepsilon}}
\Vert \theta - \theta' \Vert$. Then, we have
    \begin{align*}
         |P_n h_{\theta_0, \theta} - P_n h_{\theta_0, \xi( \theta)}|
         \leq & \left(A + M + \frac{1}{n} \sum_{i = 1}^n \Vert X_i
         \Vert \right) \varepsilon, \quad \text{and} \\ |P
         h_{\theta_0, \theta} - P h_{\theta_0, \xi( \theta)}| \leq &
         \left(A + M + \Exs \Vert X_i \Vert \right) \varepsilon \leq
         \left(A + M + \sqrt{d} \right) \varepsilon.
    \end{align*}
For any $t > 0$, as long as $\varepsilon < \frac{t}{ 3 \left(A + M +
  \sqrt{d} \right)}$, we obtain that
    \begin{align}
    \label{eq:empirical_process_first}
        \mprob \left( \sup_{\theta \in [- A, A] \times \mathcal{B}(0,
          M)} \left|P_n h_{\theta_0, \theta} - P h_{\theta_0, \theta}
        \right| \geq t \right) \leq \mprob \left( \max_{ \theta \in
          \mathcal{M}_\varepsilon} \left|P_n h_{\theta_0, \theta} - P
        h_{\theta_0, \theta} \right| \geq \frac{t}{3} \right) &
        \\ & \hspace{- 10 em} + \mprob\left( \sup_{\theta \in
          \mathcal{M}_\varepsilon} \left|P_n h_{\theta_0, \theta} -
        P_n h_{\theta_0, \xi(\theta)} \right| \geq \frac{t}{3}
        \right). \nonumber
    \end{align}
For the first term in the RHS of
equation~\eqref{eq:empirical_process_first}, we have the following
evaluation
    \begin{align*}
        \mprob \left( \max_{\theta \in \mathcal{M}_\varepsilon} \left|
        P_n h_{\theta_0, \theta} - P h_{\theta_0, \theta} \right| \geq
        \frac{t}{3} \right) & \leq |\mathcal{M}_\varepsilon|
        \max_{\theta \in \mathcal{M}_\varepsilon} \mprob \left( \left|
        P_n h_{\theta_0, \theta} - P h_{\theta_0, \theta} \right| \geq
        \frac{t}{3} \right) \\ & \leq \left( \frac{A + M}{\varepsilon}
        \right)^d \max_{\theta \in \mathcal{M}_\varepsilon} \mprob
        \left( \left| P_n h_{\theta_0, \theta} - P h_{\theta_0,
          \theta} \right| \geq \frac{t}{3} \right) \\ & \leq 2 \left(
        \frac{A + M}{\varepsilon}\right)^d \exp \left( -\frac{ nt^2}{
          72 (A+M)^2} \right).
    \end{align*}
For the second term in the RHS of
equation~\eqref{eq:empirical_process_first}, we find that
    \begin{align*}
        \mprob\left( \sup_{\theta \in \mathcal{M}_\varepsilon}
        \left|P_n h_{\theta_0, \theta} - P_n h_{\theta_0, \xi(\theta)}
        \right| \geq \frac{t}{3} \right) \leq \mprob\left( \frac{1}{n}
        \sum_{i = 1}^n \Vert X_i \Vert \geq \frac{t}{3\varepsilon}- A
        - M \right).
    \end{align*}
By choosing $\varepsilon = \frac{t}{3n(A + M + \sqrt{d \log 2 d/
    \delta})}$, the above inequality becomes
\begin{align*}
    \mprob \left( \sup_{ \theta \in \mathcal{M}_\varepsilon} 
    \left|P_n h_{\theta_0, \theta} - P_n h_{\theta_0, \xi(\theta)} \right|
    \geq \frac{t}{3} \right) \leq \delta.
\end{align*}
Getting back to $g_\theta$ from $h_{\theta,\theta_0}$, note that since
equation~\eqref{eq-equal-in-distr-emp} holds true uniformly, for any
$t>0$, we have:
\begin{align*}
    \mprob \left( \sup_{\theta \in [- A, A] \times \mathcal{B}(0, M)}
    \left|P_n g_{ \theta} - P g_{ \theta} \right| \geq t \right) &
    \leq \mprob \left( \sup_{\theta \in [- A, A] \times \mathcal{B}(0,
      M)} \left|P_n h_{\theta_0, \theta} - P h_{\theta_0, \theta}
    \right| \geq t/2 \right) \\ & + \mprob \left( \abss{
      \frac{1}{n}\sum_{i = 1}^n \Vert \xi_i \Vert^2 - d} \geq t
    \right).
\end{align*}
Standard $\chi^2$ tail bounds (Example 2.11, ~\cite{Wai19}) lead to
the following inequality
    \begin{align*}
        \mprob \left( \abss{ \frac{1}{n}\sum_{i = 1}^n
        \Vert \xi_i \Vert^2 - d} \geq t \right)
        \leq 2 \exp \left( - \frac{n t^2}{8 d} \right)
    \end{align*}
for all $t \in [0, d]$. Putting the above results together, we obtain
\begin{align*}
\mprob \left( \sup_{\theta \in [- A, A] \times \mathcal{B}(0, M)}
\left|P_n g_{ \theta} - P g_{ \theta} \right| \geq t \right) & \leq 2
\left( \frac{A + M}{\varepsilon}\right)^d \exp \left( -\frac{ nt^2}{
  288 (A+M)^2} \right) \nonumber \\ & + 2 \exp \left( - \frac{n t^2}{8
  d} \right) + \delta,
\end{align*}
where $\varepsilon = \frac{t}{3n(A + M + \sqrt{d \log 2 d/
    \delta})}$. Solving for $t$ yields
\begin{align*}
  \sup_{\theta\in [- A, A] \times \mathcal{B}(0,M)} \left|P_n
  g_{ \theta} - P g_{ \theta} \right| \leq c (1 +
  A + M) \sqrt{\frac{d}{n} \log \frac{(A + M + d) n}{\delta}},
\end{align*}
with probability at least $1 - \delta$ for some universal constant $c
> 0$. This completes the proof of the lemma.


\subsubsection{Proof of Theorem~\ref{theorem:mixing_time}}
\label{sec:proof-final}
For a scalar $s>0$, whose specific value will be determined later, we
define
\begin{align*}
A_{s} = M_{s} \defn C \left( 1 + \Vert \theta_0 \Vert + \frac{\sqrt{d
    + \log 1/ s}}{ \beta } \right),
\end{align*}
where $C$ is a positive constant in
Proposition~\ref{corollary-large-ball}.  By applying the results from
Proposition~\ref{corollary-large-ball}, we obtain that
\begin{align*}
    \pi([- A_{s}, A_{s}] \times \mathcal{B}(0, M_{s})) \geq 1 - s.
\end{align*}
For the sample size $ n \geq C_1
d\left(\beta^2(1+\Vert\theta_0\Vert^2)+d\right) \log(d (\Vert \theta_0
\Vert + 1)/ \delta)$, an application of
Lemma~\ref{lemma-empirical-process} yields that
    \begin{align*}
      |U( \theta) - U_0( \theta)|\leq \frac{1}{8},
    \end{align*}
for all $\theta \in [- A, A] \times \mathcal{B}(0, M)$ with
probability at least $1 - \delta$.  Therefore, we arrive at the
following inequalities
\begin{align*}
    e^{- 1/ 8} \int e^{- U_0( \theta)} d\theta \leq \int e^{-
      U(\theta)} d\theta \leq e^{1/ 8} \int e^{- U_0(\theta)} d\theta.
\end{align*}
The above results lead to
\begin{align}
  \label{eq:density-ratio-bound}  
  e^{- 1/ 4} \pi_0( \theta) \leq \pi( \theta) \leq e^{1/ 4}
  \pi_0( \theta).
\end{align}
Invoking the $s$-conductance result of
Proposition~\ref{cor-population-conductance}, for any set $S$ such
that $s \leq \pi_0(S) \leq \frac{1}{2}$, we obtain
\begin{align}
\label{eq:conductance-population}
\int_{S} \mathcal{T}_{x, U_0}( S^c) \pi_0(x) dx \geq C \cdot (
\pi_0(S) - s) \sqrt{\eta_{s} \cheeger_{s}},
\end{align}
for $\eta_{s} = \frac{1}{400 (A_{s} + M_{s} + \sqrt{d})^2}$ and
$\cheeger_{s} = \frac{1}{\sqrt{d}A_{s}^5M_{s}^2}$.
    
Note that for any $x$, we have $\mathcal{P}_{x, U_0} = \mathcal{P}_{x,
  U}$ since the proposal distribution is simply a spherical Gaussian
independent of the potential.  Moreover, for set $S$ such that $x
\notin S$, by the results in equation~\eqref{eq:density-ratio-bound},
we have the following evaluations
    \begin{align*}
        \mathcal{T}_{x, U}(S) = & \int_S \min \left(1, e^{U(y) - U(x)}
        \right) d \mathcal{P}_{x, U}( y) \\ \geq & \int_S \min
        \left(1, e^{U_0(y) - U_0(x)} \right) e^{- |U(x) - U_0(x)| -
          |U(y) - U_0(y)|} d \mathcal{P}_{x, U}(y) \\ \geq & e^{-
          \frac{1}{4}} \mathcal{T}_{x, U_0}(S).
    \end{align*}
Now, for any set $S \subseteq \Rspace^d$ with $2 e s \leq \pi(S) \leq
\frac{1}{2}$, the following inequalities hold
\begin{align*}
    \pi_0(S) \geq \pi_0(S \cap [- A, A]
    \times \mathcal{B}(0, M)) & \geq e^{- \frac{1}{4}}
    \pi(S \cap [- A, A] \times \mathcal{B}(0, M)) \\
    & \geq  e^{- \frac{1}{4}} (\pi(S) - s)
    \geq s.
\end{align*}
In order to study the $s$-conductance of Markov chain defined by the
RMRW algorithm, we consider two possible cases of $\pi_{0}(S)$:

\textbf{Case 1:} In the case, $\pi_0(S) \leq \frac{1}{2}$, we can
directly apply equation~\eqref{eq:density-ratio-bound} to
equation~\eqref{eq:conductance-population}, and obtain that
    \begin{align*}
        \int_{S} \mathcal{T}_{x, U} (S^{c}) \pi(x) dx
        \geq e^{- 1/ 2} \int_{S} \mathcal{T}_{x, U_0}(S^{c}) \pi_0(x) dx 
        & \geq C e^{- \frac{1}{2}} \pi_0(S) \sqrt{\eta_{s} h_{s}} \\
        & \geq C e^{- \frac{3}{4}} \pi(S) \sqrt{\eta_{s} h_{s}}.
    \end{align*}
    
\textbf{Case 2:} In the case $\pi_0(S) > \frac{1}{2}$, we have
$\pi_0(S^{c}) < \frac{1}{2}$. By the result from
equation~\eqref{eq:density-ratio-bound}, we have
\begin{align*}
  \pi_0(S) \leq e^{\frac{1}{4}} \pi(S) \leq \frac{e^{
      \frac{1}{4}}}{2}.
\end{align*}
Since the Markov chain induced by the RMRW algorithm is reversible,
the following equations hold
    \begin{align*}
        \int_{S} \mathcal{T}_{x, U_0}(S^{c}) \pi_0(x) dx = \int_S
        \int_{S^{c}} \mathcal{T}_{x, U_0}(y) \pi_0(x) dy dx & =
        \int_{S^{c}} \int_S \mathcal{T}_{y, U_0}(x) \pi_0(y) dx dy
        \\ & = \int_{S^{c}} \mathcal{T}_{y, U_0}(S) \pi_0(y) dy.
    \end{align*}
Applying the result from equation~\eqref{eq:conductance-population},
by interchanging $S$ and $S^{c}$, we obtain that
    \begin{align*}
        \int_{S^{c}} \mathcal{T}_{y, U_0}(S) \pi_0(y) dy \geq C
        \pi_0(S^{c}) \sqrt{\eta_{s} \cheeger_{s}}.
    \end{align*}
Therefore, we have the following inequalities
    \begin{align*}
        \int_{S} \mathcal{T}_{x, U}( S^{c}) \pi(x) dx \geq & e^{- 1/
          2} \int_{S^{c}} \mathcal{T}_{x, U_0}(S) \pi_0(x) dx \geq C
        e^{- 1/ 2} \pi_0(S^{c}) \sqrt{\eta_{s} \cheeger_{s}} \\ \geq&
        C e^{- 1/ 2} \frac{2 - e^{1/ 4}}{e^{1/ 4}} \pi_0(S)
        \sqrt{\eta_{s} \cheeger_{s}} \geq C \frac{2 - e^{1/ 4}}{e}
        \pi(S) \sqrt{\eta_{s} \cheeger_{s}}.
    \end{align*}
In summary, in both cases of $\pi_{0}(S)$, the $s$-conductance of
Markov chain defined by the RMRW algorithm is lower bounded by a
constant multiple of $\sqrt{\eta_{s} \cheeger_{s}}$.
    
Given the Gaussian initialization of the RMRW algorithm, we have
\begin{align*}
  \log \chi^2(\pi|| \pi^{(0)}) \lesssim d + \Vert \theta_0 \Vert^2.
\end{align*}
Therefore, the warmness is controlled by $B = e^{O(d + \Vert \theta_0
  \Vert^2)}$.  For a Markov chain with $s$-conductance $\Phi_s$, the
mixing rate theorem in the paper~\cite{lovasz2003logconcave} states
that, the total variation distance between $\pi^{(T)}$ and stationary
distribution $\pi$ is upper bounded by $B s + B e^{- T/ \Phi_s^2}$. By
choosing $s = \varepsilon/ 2 \beta$, we eventually obtain
    \begin{align*}
        T_{mix} = \frac{1}{(\sqrt{ \eta h})^2} \log \frac{B}{
          \varepsilon} \lesssim d^{1.5}(d + \Vert \theta_0
        \Vert^2)^{4.5} \log^{5.5}\frac{1}{ \varepsilon},
    \end{align*}
which completes the proof of the theorem.


\section{Discussion}
\label{sec:discussion}

In this paper, we developed and analyzed a novel polynomial mixing
time MCMC algorithm for sampling from power posterior distribution in
symmetric two-component Gaussian
mixtures~\eqref{eq:symmetric_Gaussian}.  Despite the relatively simple
structure of these models, the multi-modal nature of the posterior
poses challenges in analyzing sampling algorithms.  In order to deal
with these challenges, we introduced several new results on
Poincar\'{e} and isoperimetric inequalities for marginal and
conditional densities of power posterior distributions.

We view this work as a first step in the understanding of sampling
algorihms for mixture and hierarchical models.  First, the current
techniques in the paper are not applicable to the classical
(non-Bayesian) posterior distribution of symmetric Gaussian mixtures.
It is of interest to understand whether RMRW algorithm or other
algorithms can achieve polynomial mixing time for sampling from
classical posterior distribution in these models.  Second, the results
in the paper depend on the favorable structures of symmetric Gaussian
mixtures.  Establishing polynomial time MCMC algorithm for more
general settings of Gaussian mixtures is an interesting and important
direction.

\section*{Acknowledgments} 
\label{sec:acknowledgments}

This work was partially supported by Office of Naval Research grant
DOD ONR-N00014-18-1-2640 and National Science Foundation grant
NSF-DMS-1612948 to MJW; by NSF-DMS-grant-1909365 joint to PLB and MJW;
by Army Research Office grant W911NF-17-1-0304 to MIJ.


\appendix


\section{Proofs of various auxiliary results}
\label{subsec:remain_results}

In this appendix, we provide proofs of a number of auxiliary results
used to prove our main results.


\subsection{Proof of Lemma~\ref{lemma-quasi-concave-1d}}
\label{subsec:proof:lemma-quasi-concave-1d}

Although this lemma follows as a consequence of results from the
paper~\cite{chandrasekaran2009sampling}, we include the proof here
for the completeness.  Assume without loss of generality that $S_3$ is
a finite union of intervals, i.e., $S_3 = \bigcup_{i = 1}^N (x_i,
y_i)$.  We can further assume without loss of generality that for each
$i$, the intervals to the left of $x_i$ and to the right of $y_i$
belong to different sets (otherwise, we can remove this interval from
$S_3$). Therefore, we have
\begin{align*}
 \mathrm{dist}(S_1, S_2) = \min_{1 \leq i \leq N}(y_i - x_i).
\end{align*}
For each $i \in [N]$, we find that
\begin{align*}
  \int_{x_i}^{y_i} \pi(x) dx \geq \mathrm{dist}(S_1, S_2) \cdot
  \min(\pi(x_i), \pi(y_i)).
\end{align*}
By quasi-concavity of the function $\pi$, the sequence $(\pi( x_1),
\pi( y_1),\pi( x_2), \pi( y_2), \cdots,\pi(x_N), \pi(y_N))$ is a
concatenation of an increasing sequence and a decreasing sequence (one
of the sequences can be empty, which does not affect the proof). For
each interval in $S_1$ and $S_2$, we can assign it to an endpoint
where the value of $\pi$ is larger than that on this interval.  Two
intervals cannot be assigned to the same point, and there is at most
one interval left unassigned.  Except for the interval at the mode,
each interval in $S_1$ and $S_2$ will be assigned to the lower end of
some $(x_i, y_i)$.

Consider $\max_{1 \leq i \leq N}(\min (x_i, y_i))$. Since the interval
that achieves this maximum must have two sides with different labels,
there exists a set, say $S_1$, such that
\begin{align*}
    \max_{x \in S_1} \pi(x) \leq \max_{1 \leq i \leq N}(\min (x_i,
    y_i)).
\end{align*} 
Invoking the previous results yields
\begin{align*}
  \pi(S_1) \leq \sum_{i = 1}^N \min(\pi(x_i), \pi(y_i)) \cdot A
  \leq A \sum_{i = 1}^N \int_{ x_i}^{ y_i} \pi(x) dx \leq
  \frac{A}{\mathrm{ dist}(S_1, S_2)} \pi(S_3),
\end{align*}
which completes the proof of the lemma.


\subsection{Proof of Lemma~\ref{lemma-convex-quasi-concave}}
\label{subsec:proof:lemma-convex-quasi-concave}

Let $\varphi_m: \real^m \to \real$ be the density of $m$-dimensional
standard Gaussian distribution for $m \in \mathbb{N}_+$. We first
prove the first claim about $f_1$. Based on
equation~\eqref{eq-population-derivatives}, for any $z \in
\theta_0^{\perp}$ fixed, we derive the following equation
\begin{align*}
    \frac{\partial^2 U_0 ( a e_1 + z )}{\partial a^2} = \beta \left( 1
    - \Exs\left( \frac{ 4\varphi( X - ae_1 - z) \varphi( X + ae_1 + z
      ) }{ ( \varphi ( X - a e_1 - z ) + \varphi( X + a e_1 + z))^2}
    X_1^2 \right) \right).
\end{align*}
Taking another derivative, we find that
\begin{align*}
  \frac{ \partial^3 U_0 ( a e_1 + z )}{ \partial a^3} &
  \\ & \hspace{- 3 em} = - \Exs \frac{8 \beta X_1^3 (\varphi( X + a
    e_1 + z ) - \varphi( X - a e_1 - z )) \varphi( X + a e_1 + z )
    \varphi( X - a e_1 - z )}{(\varphi( X + ae_1 + z ) + \varphi( X
    - ae_1 - z ))^3 }.
\end{align*}

Note that the distribution of $X$ in the above expectation is
symmetric around zero.  Moreover, we have $\varphi( X - a e_1 - z) =
\varphi_1( X_1 - a ) \varphi_{d-1} ( X_{-1} - z )$ and $\varphi( X +
ae_1 + z) = \varphi_1( X_1 + a ) \varphi_{d-1} ( X_{-1} + z )$.  Using
these facts, some simple algebra leads to
\begin{align*}
  \frac{ \partial^3 U_0 ( ae_1 + z )}{ \partial a^3} & \\
  & \hspace{- 6 em} =  8 \beta \Exs \left( \frac{ X_1^3 \left(  e^{a X_1} 
    \varphi_{d - 1} ( X_{- 1} - z ) -  e^{- a X_1}
    \varphi_{d - 1} ( X_{- 1} + z ) \right)  \varphi_{d-1} 
    ( X_{- 1} + z )  \varphi_{d - 1} ( X_{- 1} - z )}
       {( e^{a X_1} \varphi_{d - 1} ( X_{- 1} - z ) 
    +  e^{- a X_1} \varphi_{d - 1} ( X_{- 1} + z ))^3 }\right).
\end{align*}
For $a > 0$, since $X_1^3, e^{a X_1} - 1, 1 - e^{- a X_1}$ always have
the same signs, we find that
\begin{align*}
    \frac{ \partial^3 U_0 ( a e_1 + z )}{ \partial a^3} \\ & \hspace{-
      6 em} = 8 \beta \Exs \left( \frac{ X_1^3 \left( e^{a X_1}
      \varphi_{d - 1} ( X_{- 1} - z ) - e^{-a X_1} \varphi_{d - 1} (
      X_{- 1} + z ) \right) \varphi_{d - 1} ( X_{- 1} + z ) \varphi_{d
        - 1} ( X_{- 1} - z )}{( e^{a X_1} \varphi_{d - 1} ( X_{- 1} -
      z ) + e^{-aX_1} \varphi_{d - 1} ( X_{-1} + z ))^3 } \right)
    \\ & \hspace{- 6 em} \geq 8 \beta \Exs \left( \frac{ X_1^3 \left(
      \varphi_{d - 1} ( X_{- 1} - z ) - \varphi_{d - 1} ( X_{- 1} + z
      ) \right) \varphi_{d - 1} ( X_{-1} + z ) \varphi_{d - 1} ( X_{-
        1} - z )}{( e^{a X_1} \varphi_{d - 1} ( X_{- 1} - z ) + e^{- a
        X_1} \varphi_{d - 1} ( X_{- 1} + z ))^3 } \right) = 0.
\end{align*}
The last equation in the above display is due to the symmetry of
distribution of $X$.  Therefore, $\frac{ \partial U_0(a e_1 +
  z)}{\partial a}$ is a convex function of $a \in
\real_+$. Furthermore, the following identities can be derived by a
combination of algebra and integral transforms:
    \begin{align*}
    	\frac{ \partial U_0(a e_1 + z)}{ \partial a} \Big|_{a = 0} = &
        a_0 - 2 a_0 \Exs \parenth{\frac{1}{1 + \exp(4 z^{\top} \xi_{ -
              1})}} = 0, \quad \text{and} \\ \frac{ \partial U_0( a
          e_1 + z)}{\partial a} \Big|_{a = a_0} = & - 2 \Exs
        \left(\frac{a_0 + \xi_1}{1 + \exp(2 a_0 (a_0 + \xi_1) + 4
          z^{\top} \xi_{ - 1})}\right) = 0.
    \end{align*}
Given the above results, we have $\frac{\partial U_0(a e_1 +
  z)}{\partial a} \leq 0$ on the interval $[0, a_0]$, and $\frac{
  \partial U_0(a e_1 + z)}{ \partial a} \geq 0$ for $a \geq a_0$. As a
consequence, we achieve the conclusion of the lemma.

Now we turn to prove the second claim about $f_2$. Given $a > 0$
fixed, using equation~\eqref{eq-population-derivatives}, we obtain
that
\begin{align*}
  \nabla_z^2 U_0(a e_1 + z) = \beta I_{d} - \beta \Exs
  \parenth{ \frac{ 4 \varphi(X - \theta) \varphi(X +
      \theta)} {(\varphi(X - \theta) + \varphi(X +
      \theta))^2} \xi_{- 1} \xi_{- 1}^{\top}}.
\end{align*}
Thus, we have $\xi_{- 1} \xi_{- 1}^{\top} \succeq 0$, and $4 \varphi(X
- \theta) \varphi(X + \theta) \leq (\varphi(X - \theta) + \varphi(X +
\theta))^2$. These inequalities lead to
\begin{align*}
  \nabla_z^2 U_0(a e_1 + z) \succeq \beta( I_{d} -
  \Exs(\xi_{- 1} \xi_{-1}^{\top}) ) = 0,
\end{align*}
as claimed.


\subsection{Proof of Lemma~\ref{lemma-dissipative-tail}}
\label{subsec:proof:lemma-dissipative-tail}

Consider the Langevin diffusion defined by the following It\^{o}
stochastic differential equation (SDE):
\begin{align}
\label{eq:langevin-diffusion}  
    dX_t = - \nabla \bar{ U} ( X_t) dt + \sqrt{2} dB_t,
  \end{align}
with initial condition $X_0 = 0$. It is known that under the
dissipativity assumption assumed in this lemma, the distribution of
the diffusion process $X_t$ converges to the stationary distribution
$\pi$ as $t$ tends to infinity~\cite{risken1996fokker}.  Our next step
is to derive an upper bound for $\Exs \Vert X_t \Vert^p$ with $p \geq
1$.
  
Invoking It\^{o}'s formula, for any $\alpha > 0$, we have
\begin{align*}
  \frac{1}{2} e^{ \alpha t} \Vert X_t \Vert^2 - \frac{1}{2} \Vert X_0
  \Vert^2 & = \int_0^t \langle X_s, - \nabla \bar{ U} ( X_s) e^{
    \alpha s} \rangle ds + \frac{d}{2} \int_0^t e^{\alpha s} ds \\ & +
  \int_0^t e^{ \alpha s} X_s^{\top} dB_s + \frac{1}{2} \int_0^t c
  e^{\alpha s} \Vert X_s \Vert^2 ds.
\end{align*}
Let $M_t \mydefn \int_0^t X_s^{\top} e^{ \alpha s} dB_s$ be the
martingale term.  Without loss of generality, we can assume that $p
\geq 4$; the case $p \in [1,4]$ can be derived as a direct consequence
of the result $p \geq 4$ by applying the H\"{o}lder inequality.
Applying the Burkholder-Gundy-Davis inequality yields
\begin{align*}
  \Exs \sup_{0 \leq t \leq T} |M_t|^{ \frac{p}{2}} \leq (p C)^{
    \frac{p}{4}} \Exs \langle M, M \rangle_T^{ \frac{p}{4}} = & (p
  C)^{ \frac{p}{4}} \Exs \left( \int_0^T e^{ 2 \alpha s} X_s^{ \top}
  X_s ds \right)^{ \frac{p}{4}} \\ \leq & (p C)^{ \frac{p}{4}} \Exs
  \left( \int_0^T e^{ 2 \alpha s} \Vert X_s \Vert^2 ds \right)^{
    \frac{p}{4}} \\ \leq & (p C)^{ \frac{p}{4}} \Exs \left( \sup_{0
    \leq s \leq T} e^{ \alpha s} \Vert X_s \Vert^2 \cdot \int_0^T e^{
    \alpha s} ds \right)^{ \frac{p}{4}} \\ \leq & \left( \frac{ C p
    e^{\alpha T}}{\alpha} \right)^{ \frac{p}{4}} \left(A + \frac{1}{A}
  \Exs \left( \sup_{0 \leq t \leq T} e^{ \alpha t} \Vert X_t \Vert^2
  \right)^{ \frac{p}{2}} \right),
\end{align*}
where $C > 0$ is some universal constant and $A$ is an arbitrary
number which will be determined later. On the other hand, by the
dissipativity assumption in the lemma, we have
\begin{align*}
  \int_0^t \langle X_s, -\nabla \bar{ U} ( X_s) e^{ \alpha s}
  \rangle ds \leq \int_0^t \left( - a \Vert X _s \Vert^2 + b
  \right) e^{ \alpha s} ds.
\end{align*}
Putting the above results together and letting $\alpha = 2 a$, we
obtain that
\begin{align*}
  \Exs \left( \sup_{0 \leq t \leq T} e^{2 a t} \Vert X_t \Vert^2
  \right)^{ \frac{p}{2}} & \leq 3^{ \frac{p}{2} - 1} \Exs \biggr(
  \sup_{0 \leq t \leq T} \int_0^t \biggr( 2 \langle X _s, - \nabla
  \bar{U} ( X _s) \rangle + d + c \Vert X_s \Vert^2 \biggr) e^{cs} ds
  \biggr)^{ \frac{p}{2}} \\ & \hspace{ 12 em} + 3^{ \frac{p}{2} - 1}
  \Exs \sup_{0 \leq t \leq T} |M_t|^{ \frac{p}{2}} \\ & \leq 3^{
    \frac{p}{2} - 1} \left( \frac{C p e^{2 a T}}{2 a} \right)^{
    \frac{p}{4}} \left( A + \frac{1}{A} \Exs \left( \sup_{0 \leq t
    \leq T} e^{2 a t} \Vert X_t \Vert^2 \right)^{ \frac{p}{2}} \right)
  \\ & \hspace{ 12 em} + 3^{ \frac{p}{2} - 1} \Exs \left( \sup_{0 \leq
    t \leq T} \int_0^t \left(2 b + d \right) e^{2 a s} ds \right)^{
    \frac{p}{2}},
\end{align*}
for some universal constant $C>0$.

Setting $A \defn 2 \cdot 3^{ \frac{p}{2} - 1} \left( \frac{C p e^{ 2 a
    T}}{a} \right)^{ \frac{p}{4}}$ in the above inequality, we find
that
\begin{align}
  \left( \Exs \Vert X_T \Vert^p \right)^{ \frac{1}{p}} \leq e^{- a T}
  \left( \Exs \left( \sup_{0 \leq t \leq T} e^{ 2 a t} \Vert X_t
  \Vert^2 \right)^{ \frac{p}{2}} \right)^{ \frac{1}{p}} \leq C' \left(
  \sqrt{ \frac{p}{a}} + \sqrt{ \frac{b + d}{a}} \right),
\end{align}
for some universal constant $C' > 0$. Letting $T \to +\infty$ yields
\begin{align*}
  (\Exs_\pi \Vert X \Vert^p )^{ \frac{1}{p}}
  \leq C' \left(
  \sqrt{ \frac{p}{a}} + \sqrt{ \frac{b + d}{a}} \right).
\end{align*}
Furthermore, for any $t > 0$, an application of Markov's inequality leads to 
\begin{align*}
  \mprob \left(\Vert X \Vert \geq t \right) \leq \inf_{p \geq 1}
  \frac{ \Exs \Vert X \Vert^p}{ t^p} \leq \inf_{p \geq 1} (2C')^p
  \left( \left( \frac{p}{ a t^2} \right)^{ \frac{p}{2}} + \left(
  \frac{b + d}{ a t^2} \right)^{ \frac{p}{2}} \right).
\end{align*}
For any $\delta \in (0,1)$, setting $p = 2 \log \frac{2}{\delta}$ and
$t = 2 C'( \sqrt{ \frac{p}{a}} + \sqrt{\frac{b + d}{a}})$ yields
\begin{align*}
  \mprob \left( \Vert X \Vert \geq t \right) \leq \left( \frac{ 4 C'^2
    p}{a t^2} \right)^{ \frac{p}{2}} + \left( \frac{ 4 C'^2 (b + d)}{a
    t^2} \right)^{ \frac{p}{2}} \leq \frac{\delta}{2} + \frac{
    \delta}{2} = \delta,
\end{align*}
which completes the proof of the lemma.

\subsection{Proof of Lemma~\ref{lemma-rejection-overlap}}
\label{subsec:lemma:lemma-rejection-overlap}

At the population level, we always have $U_{0}(Z) = U_{0}(Y)$ where
$Y$ and $Z$ are given in the RMRW algorithm
(Algorithm~\ref{alg-MCMC-sym}).  Therefore, the rejection step can
only be caused by the difference between $F(Y)$ and $F(\theta^{(t -
  1)})$ where $\theta^{(t)}$ are samples from the RMRW algorithm.  By
Gaussian tail bounds, we have
\begin{align*}
  \mathcal{P}_{x, U_0} \left( \mathcal{B}(x, 10 \sqrt{d \eta})^{c}
  \right) \leq \frac{1}{20}.
\end{align*}
Therefore, we can restrict our attention inside the ball
$\mathcal{B}(x, 10 \sqrt{d \eta})$.  Now, for $y \in \mathcal{B}(x, 5
\sqrt{d \eta})$ and $x \in [- A, A] \times \mathcal{B}(0, M)$, we find
that
\begin{align*}
  |U_{0}(x) - U_{0}(y)| \leq \Vert x - y \Vert \sup_{x \in [- A, A]
    \times \mathcal{B}(0, M)} \Vert \nabla U_0(x) \Vert \leq 2(A + M +
  \sqrt{d}) \Vert x - y \Vert.
\end{align*}
By choosing $\eta \in \big(0, \frac{1}{400(A+M+\sqrt{d})^2} \big)$, we
can ensure that
\begin{align*}
  \min(1, \exp(F(y) - F(x))) \geq 1 - \frac{1}{20}.
\end{align*}
Combining the previous inequalities together yields the first
inequality of the lemma.
    
For the overlap bound, standard application of triangle inequality
with the total variation distance yields that
    \begin{align*}
      \dtv(\mathcal{T}_{x, U_0}, \mathcal{T}_{y, U_0}) \leq \dtv
      (\mathcal{T}_{x, U_0}, \mathcal{P}_{x, U_0}) + \dtv
      (\mathcal{P}_{x, U_0}, \mathcal{P}_{y, U_0}) + \dtv
      (\mathcal{P}_{y, U_0}, \mathcal{T}_{y, U_0}),
    \end{align*}
In order to prove the second bound of the lemma, it is sufficient to
bound the term $\dtv( \mathcal{P}_{x,U_0}, \mathcal{P}_{y,U_0})$.  For
the case $\max \left\{\Vert x-y\Vert,\Vert x+y\Vert\right\} \leq
\frac{1}{4}\eta$, we assume without loss of generality that $\Vert x -
y \Vert \leq \frac{1}{10} \sqrt{\eta}$.  Then, the following
inequalities hold
\begin{align*}
  \dtv( \mathcal{P}_{x, U_0}, \mathcal{P}_{y, U_0}) & \leq \frac{1}{2}
  \dtv (\mathcal{N}(x, \eta I_{d}), \mathcal{N}(y,\eta I_{d})) +
  \frac{1}{2} \dtv (\mathcal{N}( - x, \eta I_{d}), \mathcal{N}(- y,
  \eta I_{d})) \\
& \leq \frac{1}{2} \sqrt{\text{KL} (\mathcal{N} (x, \eta I_{d}),
    \mathcal{N} (y, \eta I_{d}))} + \frac{1}{2}
  \sqrt{\text{KL}(\mathcal{N}(- x, \eta I_{d}), \mathcal{N}(- y,\eta
    I_{d}))} \\
& \leq \frac{1}{10}.
\end{align*}
Putting the above inequalities together yields the second bound in the
lemma statement.


\subsection{Proof of Proposition~\ref{prop-robust-misspecify}}
\label{subsec:proof:prop-robust-misspecify}

The proof of the proposition hinges upon several arguments similar to
those used in the proofs for correctly-specified model.  Therefore, we
provide only detailed proofs for steps that are fundamentally
different between these settings.

By Assumption~\ref{assume-robustness}, the true distribution can be
written as $Q = (1 - \gamma) P_{0} + \gamma F$, with $P_{0} =
\frac{1}{2} \mathcal{N} ( \theta_0, I_{d}) + \frac{1}{2} \mathcal{N}(-
\theta_0, I_{d})$ and $F$ being an arbitrary $K$-sub-Gaussian
distribution.  While $U_{0}$ is not the exact population power
posterior distribution in the contaminated model, it is nonetheless
very close to the true population power posterior because $\gamma$ is
sufficiently small. Thus, this approximation can be used for the proof
of the proposition.

Throughout the remainder of this proof, we use the isoperimetric
inequalities for ``population'' power posterior defined by $P_0$,
which depends on neither the contaminated distribution nor the data.
There are essentially two places in the proofs of mixing rate of the
RMRW algorithm that rely on the distribution of samples:
Corollary~\ref{corollary-large-ball} and
Lemma~\ref{lemma-empirical-process}.  Therefore, we will adjust these
results to contaminated models separately.

For the proof of Corollary~\ref{corollary-large-ball} under the
correctly-specified model, we need to establish a lower bound on the
term $\langle \nabla U(\theta) , \theta \rangle$. For the given
mis-specified model, the following lemma gives a lower bound for this
term:
\begin{lemma}
\label{lemma-robust-large-ball}
Under Assumption~\ref{assume-robustness}, with probability at least $1
- \delta$, for the sample size $n \geq c (d + \Vert \theta_0 \Vert^2)
\log \frac{ d + \Vert \theta_0 \Vert^2 }{\delta}$ for some universal
constant $c > 0$, we have
\begin{align*}
  \langle \nabla U(\theta), \theta \rangle \geq \frac{\beta}{2} \Vert
  \theta \Vert^2 - 2 \beta \left( \Vert\theta_0 \Vert^2 + 1 + \gamma d
  K^2 \log \frac{n}{\delta} \right)
\end{align*}
for any $\theta \in \real^d$.
\end{lemma}
\noindent See Appendix~\ref{subsec:proof:lemma-robust-large-ball} for
the proof of Lemma~\ref{lemma-robust-large-ball}.

Similarly, an adaptation of Lemma~\ref{lemma-empirical-process} from
the well-specified case to the mis-specified case leads to the
following result:
\begin{lemma}
\label{lemma-robust-empirical-process}
Under Assumption~\ref{assume-robustness}, for fixed constants $A , M
\geq \Vert \theta_0 \Vert + 1$, with probability at least $1 -
\delta$, we have
\begin{align*}
  \sup_{\theta \in [- A, A] \times \mathcal{B}(0, M)}
  \left|\frac{1}{n} \sum_{ i = 1}^n g_{\theta} (X_i) - \Exs_{P_{0}}
  g_{\theta} (X) \right| \lesssim (1 + A + M) \sqrt{ \frac{d}{n} \log
    \frac{n ( A + M + d)}{\delta}} & \\ & \hspace{- 8 em} + \gamma(
  A^2 + M^2 + K^2 d \log \frac{n}{\delta}).
\end{align*}
\end{lemma}
\noindent
See Appendix~\ref{subsec:proof:lemma-robust-empirical-process} for the
proof of this claim. \\

With these two lemmas at our disposal, we are ready to prove
Proposition~\ref{prop-robust-misspecify}.  The geometric results of
population power posterior under $P_0$ for correctly-specified case
are still valid for the mis-specified setting.  In order to use the
$s$-conductance framework for the Markov chain with $U$, we need to
control the probability of a region outside a large ball, which is
done in Lemma~\ref{lemma-robust-large-ball} with radius $A_s = M_s =
\Vert \theta_0 \Vert + 1 +K \sqrt{ \gamma d \log \frac{n}{\delta} } +
\beta^{- 1} \sqrt{d + \log 1/s}$.  According to
Section~\ref{sec:proof-final}, we only need to establish a uniform
upper bound on the difference between $U_0$ and $U$ of constant order.
This has been done in Lemma~\ref{lemma-robust-empirical-process},
under the condition on $\gamma$.  As a consequence, we reach the
conclusion of the proposition.


\subsubsection{Proof of Lemma~\ref{lemma-robust-large-ball}}
\label{subsec:proof:lemma-robust-large-ball}

For each $i \in [n]$, $X_i$ can be written as $X_i = (1 - \xi_i) Y_i +
\xi_i E_i$, where $Y_i \sim P_{0}$, $E_i\sim F$ and $\xi_i \sim
\mathrm{Bern}(\gamma)$ are independent random variables. Invoking
Chernoff's bound, we find that
\begin{align*}
    \mprob\left( \sum_{i = 1}^n \xi_i \geq 2 \eta n \right) \leq e^{-
      n D_{KL}(2\eta||\eta)} \leq e^{- (2 \log 2 - 1) n}.
\end{align*}
Therefore, for $n \geq C \log \delta^{-1}$, we have $\sum_{i = 1}^n
\xi_i \leq 2 \eta n$ with probability at least $1 - \delta/ 2$. Recall
that $L(\theta; X_{i}) = \frac{1}{2} \varphi( \theta - X_i) +
\frac{1}{2} \varphi(\theta + X_i)$ for all $i \in [n]$.  Conditioning
on the previous event, we obtain that
\begin{align*}
    \langle \nabla U (\theta), \theta \rangle = & \frac{1}{n} \sum_{i:
      \xi_i = 0} \langle \nabla \log L(\theta; X_i) , \theta \rangle +
    \frac{1}{n} \sum_{i:\xi_i = 1} \langle \nabla \log L(\theta; X_i)
    , \theta \rangle \\ \stackrel{(i)}{ \geq} & ( 1 - 2\gamma ) \left(
    \frac{\beta}{2} \Vert \theta \Vert^2 - 2\beta (\Vert\theta_0
    \Vert^2 +1 ) \right) + 2\gamma \beta \frac{1}{n} \sum_{i:\xi_i =
      1} \left( \Vert \theta \Vert^2 - \Vert \theta \Vert \cdot \Vert
    X_i \Vert \right)\\ \stackrel{(ii)}{ \geq} & \frac{\beta}{2} \Vert
    \theta \Vert^2 - 2 \beta \left( \Vert\theta_0 \Vert^2 + 1 + \gamma
    d K^2 \log \frac{n}{\delta} \right),
\end{align*}
with probability $1 - \delta/ 2$. In the above display, inequality (i)
follows from the proof of Proposition~\ref{corollary-large-ball}, and
inequality (ii) is a direct consequence of the $K$-sub-Gaussian
assumption on $F$. As a consequence, we obtain the conclusion of the
lemma.


\subsubsection{Proof of Lemma~\ref{lemma-robust-empirical-process}}
\label{subsec:proof:lemma-robust-empirical-process}

We adopt the same notation as in the proof of
Lemma~\ref{lemma-robust-large-ball}.  Note that we have the bound
$\frac{1}{\obs} \sum_{i = 1}^\obs \xi_i \leq 2 \eta$ with probability
$1 - \delta/ 2$.  Conditioning on this event, an application of the
triangle inequality yields
\begin{align*}
\left|\frac{1}{\obs} \sum_{i=1}^\obs g_{\theta} (X_i) - \Exs_{P_{0}}
g_{\theta} (X) \right| \leq \left| \frac{1}{\obs} \sum_{i:\xi_i = 0}
g_{\theta} (X_i) - \Exs_{P_{0}} g_{\theta} (X) \right| +
\left|\frac{1}{\obs} \sum_{i:\xi_i = 1} g_{\theta} (X_i) -
\Exs_{P_{0}} g_{\theta} (X) \right|.
\end{align*}
The supremum of the first term is controlled by
Lemma~\ref{lemma-empirical-process}. The second term can be bounded as
follows:
\begin{align*}
  \sup_{\theta \in [- A, A] \times \mathcal{B}(0, M)}
  \left|\frac{1}{n}\sum_{i:\xi_i = 1} g_{\theta} (X_i) - \Exs_{P_{0}}
  g_{\theta} (X) \right| \leq & 2 \sup_{\theta \in [- A, A] \times
    \mathcal{B}(0, M)} \frac{1}{n} \sum_{i: \xi_i = 1} \left(\Vert
  \theta \Vert^2 + \Vert X_i\Vert^2 \right) \\ \leq & 4 \gamma( A^2 +
  M^2 + K^2 d \log \frac{n}{\delta}),
\end{align*}
a bound that holds with probability at least $1 - \delta/ 2$.
Combining the above results yields the claim in the lemma.


\bibliography{Nhat}


\end{document}